\documentclass[dvipsnames,format=sigconf,authorversion=true,nonacm]{acmart}

\usepackage[ruled,noend,linesnumbered,noline]{algorithm2e}
\SetFuncSty{textsc}
\usepackage{hyperref}
\usepackage[noabbrev,capitalise]{cleveref}
\usepackage{array}
\usepackage{booktabs}
\usepackage{tabularx}
\usepackage{makecell}
\usepackage{multirow}
\usepackage{multicol}
\usepackage{subcaption}
\def\vec#1{\mathchoice{\mbox{\boldmath$\displaystyle#1$}}
{\mbox{\boldmath$\textstyle#1$}}
{\mbox{\boldmath$\scriptstyle#1$}}
{\mbox{\boldmath$\scriptscriptstyle#1$}}}

\setcopyright{cc}

\acmDOI{10.1145/nnnnnnn.nnnnnnn} 
\acmISBN{978-x-xxxx-xxxx-x/YY/MM} 
\acmConference[GECCO '26]{The Genetic and Evolutionary Computation Conference 2026}{July 13--17, 2026}{San Jos\'e, Costa Rica} 
\acmYear{2026}
\copyrightyear{2026}

\setcctype{by-nc-nd}
\begin{document}

\title{Introns and Templates Matter: Rethinking Linkage in GP‑GOMEA}

\author{Johannes Koch}
\orcid{0009-0008-7570-7621}
\affiliation{%
  \institution{Centrum Wiskunde \& Informatica}
  \city{Amsterdam}
  \country{The Netherlands}
}
\affiliation{%
  \institution{Delft University of Technology}
  \city{Delft}
  \country{The Netherlands}
}
\email{johannes.koch@cwi.nl}

\author{Tanja Alderliesten}
\orcid{0000-0003-4261-7511}
\affiliation{%
  \institution{Leiden University Medical Center}
  \city{Leiden}
  \country{The Netherlands}
}
\email{T.Alderliesten@lumc.nl}

\author{Peter A.N. Bosman}
\orcid{0000-0002-4186-6666}
\affiliation{%
  \institution{Centrum Wiskunde \& Informatica}
  \city{Amsterdam}
  \country{The Netherlands}
}
\affiliation{%
  \institution{Delft University of Technology}
  \city{Delft}
  \country{The Netherlands}
}
\email{peter.bosman@cwi.nl}

\renewcommand{\shortauthors}{Koch et al.}

\begin{abstract}
GP-GOMEA is among the state-of-the-art for symbolic regression, especially when
it comes to finding small and potentially interpretable solutions. A key
mechanism employed in any GOMEA variant is the exploitation of linkage, the
dependencies between variables, to ensure efficient evolution. In GP-GOMEA,
mutual information between node positions in GP trees has so far been used
to learn linkage. For
this, a fixed expression template is used. This however leads to introns for
expressions smaller than the full template. As introns have no impact on
fitness, their occurrences are not directly linked to selection. Consequently,
introns can adversely affect the extent to which mutual information captures
dependencies between tree nodes. To overcome this, we propose two new
measures for linkage
learning, one that explicitly considers introns in mutual
information estimates, and one that revisits linkage learning in GP-GOMEA from a
grey-box perspective, yielding a measure that needs not to be learned from
the population but is derived directly from the template. Across five standard
symbolic regression problems, GP-GOMEA achieves substantial improvements using
both measures. We also find that the newly learned linkage structure closely
reflects the template linkage structure, and that explicitly using the template
structure yields the best performance overall.
\end{abstract}




\maketitle

\section{Introduction}

Especially in domains where decisions can have severe consequences on lives and livelihoods, such as healthcare, accountability and responsibility are paramount. In recognition of this, the fields of eXplainable AI (XAI) and interpretable machine learning aim to either explain model outputs or learn models that are inherently transparent~\cite{rudinStopExplainingBlack2019, virgolinLessMoreCall2022a}. Symbolic regression (SR), the task of finding mathematical expressions that accurately model relationships in data, is one form of such inherently interpretable models~\cite{kozaGeneticProgrammingMeans1994a}. Small mathematical expressions describing one target variable consisting of atomic functions (e.g., \(\{+,-,\times,\div,\sin\}\)), numerical parameters, and input features potentially are not only readable for human experts, but also have the potential to be interpretable and can help uncover yet unknown knowledge.

GP-GOMEA, a Genetic Programming (GP) version of the Gene-pool Optimal Mixing Evolutionary Algorithm (GOMEA), is one of the leading algorithms when it comes to finding small, yet accurate symbolic expressions~\cite{aldeiaCallActionNext2025a,lacavaContemporarySymbolicRegression2021}. GP algorithms typically start with an initially random population of solutions that is refined through repeated selection and variation. One key aspect to GP-GOMEA is the explicit modeling and exploitation of key dependencies between decision variables, also called linkage~\cite{virgolinImprovingModelbasedGenetic2021a}. Linkage is well-known to be a crucial factor in how effectively an algorithm can tackle problems in general, where problems can be exponentially harder to solve without the correct linkage structure~\cite{thierensScalabilityProblemsSimple1999}. If no linkage information is known a priori, it can be detected and learned during optimization. Oftentimes, this happens based on statistics taken from the current (and potentially previous) population(s). Specifically, a measure is needed that expresses how strongly variables are related. For this, it was previously proposed to use mutual information (MI) between decision variables in GP-GOMEA, with a modification to avoid the initial detection of false linkage due to non-uniform initialization methods~\cite{virgolinImprovingModelbasedGenetic2021a}.

Similarly to other approaches that explicitly maintain distributions over the expressions in the search space~\cite{salustowiczProbabilisticIncrementalProgram1997, poliLinearEstimationofDistributionGP2008,hembergInvestigationLocalPatterns2012}, a fixed template is used to ensure all solutions have the same number of variables and that variable indices correspond to identical expression positions across solutions. Oftentimes, without further knowledge of the type of expression being searched for, the template corresponds to a full binary tree with a fixed depth. For expressions that do not use the entire template, some decision variables are conditionally inactive, often called introns~\cite{loboCompressedIntronsLinkage}.

However, from the perspective of linkage learning, introns can be problematic, as the absense of selection pressure on introns causes them to act as a source of noise to the statistics used to learn linkage. Inactive variables even have been recognized as a source of difficulty beyond GP~\cite{guijtExploringSearchSpace2024,guijtStitchingNeuroevolutionRecombining2024,przewozniczekHoplikeProblemNature2024,przewozniczekConditionalDirectEmpirical2025}.

The goal of this paper is to study whether explicitly taking the presence of introns into account during linkage learning can improve the detection of important dependencies and consequently lead to improved performance by GP-GOMEA. Moreover, we wish to study closer what dependencies are discovered once the noise caused by the presence of introns is removed.

The remainder of this paper is organized as follows: Related work is covered in~\Cref{sec:related_work}. GP-GOMEA and new ways of detecting linkage are discussed in~\Cref{sec:method}. Experiments and results thereof are covered in~\Cref{sec:experiments}. Finally, we discuss our results in~\Cref{sec:discussion} and conclude in~\Cref{sec:conclusion}.

\section{Related Work\label{sec:related_work}}

The importance of linkage in evolutionary optimization is well-known~\cite{chenSurveyLinkageLearning2007} and it has been shown that the computational effort needed to solve optimization problems can exponentially increase without the right linkage model~\cite{thierensScalabilityProblemsSimple1999}. There are various ways to handle linkage to more effectively propagate building blocks with less disruption: The solution representation can be optimized towards highly linked low-order schemata~\cite{hollandAdaptationNaturalArtificial1992}, variables can be re-ordered during evolution to improve linkage~\cite{loboCompressedIntronsLinkage,goldbergMessyGeneticAlgorithms}, domain specific crossovers can be used~\cite{quevedodecarvalhoDramaticallyFasterPartition2025}, distribution-based algorithms can encode linkage in probabilistic models~\cite{pelikanBOABayesianOptimization1999,bosmanLinkageInformationProcessing1999}, or linkage can be modeled and learned explicitly. GOMEAs fall into the last category of model-based evolutionary algorithms, where linkage information is either known or learned~\cite{dushatskiyParameterlessGenepoolOptimal2021a, przewozniczekConditionalDirectEmpirical2025,andreadisFitnessbasedLinkageLearning2024}.

In the field of GP, too, decision variables are typically arranged such that related variables are grouped and varied together, for example by representing solutions as trees in standard GP. All the above-mentioned approaches arguably have been explored to various extents, albeit not necessarily with an explicit intent towards linkage in GP: Tree-based GP keeps related variables close together~\cite{kozaGeneticProgrammingMeans1994a}, linear and cartesian GP allow re-ordering subroutines by decoupling semantic from spatial adjacency~\cite{wilsonComparisonCartesianGenetic2008}, various context aware crossover operators~\cite{majeedLessDestructiveContextAware2006} and distribution based GP algorithms~\cite{salustowiczProbabilisticIncrementalProgram1997,poliLinearEstimationofDistributionGP2008,songSymbolGraphGenetic2024a} have been proposed. How linkage between variables is modeled and exploited is generally highly specific to the solution representation and algorithm used.

When it comes to explicitly modeling and learning linkage as done in GOMEA, various topics have been explored. From different ways of modeling and learning linkage, to similarity measures, to filtering extraneous linkage, to conditional variation operators that consider the context of already varied linked variables~\cite{bosmanMeasuresBuildLinkage2012,bosmanLinkageNeighborsOptimal2012,bosmanMoreConciseRobust2013,andreadisFitnessbasedLinkageLearning2024,przewozniczekDirectLinkageDiscovery2021,przewozniczekConditionalDirectEmpirical2025,dushatskiyParameterlessGenepoolOptimal2021a}. In particular, fitness-based linkage learning, i.e., performing targeted fitness evaluations to uncover definitive linkage instead of using potentially untruthful statistical measures, is gaining ground~\cite{przewozniczekDirectLinkageDiscovery2021,przewozniczekConditionalDirectEmpirical2025,andreadisFitnessbasedLinkageLearning2024}. Due to the use of a fixed template in GP-GOMEA and the knowledge that it is possible for any two variables to interact, this form of linkage learning does not apply to symbolic regression (SR). 
The linkage measure used typically heavily depends on the domain~\cite{bosmanExpandingDiscreteCartesian2016,bouterExploitingLinkageInformation2017a,sadowskiLearningExploitingMixed2016}.

To the best of our knowledge, additional information available in the GP setting such as the user-defined template structure or knowledge about inactive variables has not yet been considered when it comes to linkage. Nonetheless, an adjustment to the estimated linkage to prevent the detection of spurious linkage after initialization and a binning approach to handle numerical constants were previously proposed~\cite{virgolinImprovingModelbasedGenetic2021a}. This adjusted linkage measure is re-visited and compared to in this work.
\section{Method\label{sec:method}}

In this section, GP-GOMEA is introduced, before the proposed changes to how linkage is learned, are presented.

\subsection{GP-GOMEA}

As in other population-based algorithms, GP-GOMEA works by iteratively improving an initial, often random, set of candidate solutions through variation and selection. Compared to most other algorithms, a linkage learning step is performed every generation in GP-GOMEA, followed by variation using the the gene-pool optimal mixing (GOM) operator. GOM combines both variation and selection on a per-solution basis as shown in~\Cref{code:gpgomea}.

\begin{figure}[htbp]
    \centering
    \vspace{-0.2cm}
    \includegraphics[width=0.9\linewidth]{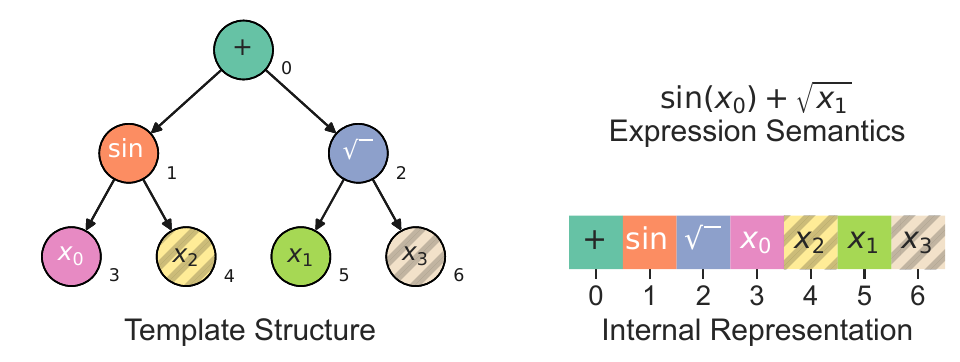}
    \Description{A figure showing the internal representation corresponding to the expression $\sin(x_0)+\sqrt{x_1}$, and how this representation is mapped to an expression tree with the use of a shared template. The template in the example is a binary tree of depth 2, where each variable index in the internal representation maps to a node position in the shared structural template, leading to the final expression. As unary functions like sine and square root do not use both children, the variables corresponding to those template positions in the example expression are inactive.}
    \vspace*{-3mm}
    \caption{A fixed-length string of decision variables is mapped to a fixed tree template, which defines the corresponding semantic expression. Shaded variables are introns, and do not affect the semantic meaning of the expression.}
    \label{fig:template}
    \vspace{-0.1cm}
\end{figure}

Each solution consists of a fixed-length string of decision variables, where a common tree-based template structure is used to map decision variables to tree nodes, as shown in~\Cref{fig:template}. Typically, this template is a single $n$-ary tree, where $n$ is the maximum arity in the function set used~\cite{virgolinImprovingModelbasedGenetic2021a}. However, the use of multiple output trees or modular subfunctions is also possible~\cite{sijbenMultimodalMultiobjectiveModelbased2022a,harrisonThinkingOutsideTemplate2025}. By using a fixed structural template compared to a representation with a dynamic length, the same variable position corresponds to the same tree node across all individuals. Hence, the linkage between variables corresponds to the linkage of the respective tree nodes. However, this inherently introduces conditionally inactive variables or \emph{introns}, as for smaller expressions not all decision variables are used in the corresponding semantic expression.

In each generation, first a linkage model is learned, aimed at capturing the dependencies between variables. To represent this linkage information, a set containing multiple subsets of all decision variables, called a Family of Subsets (FOS, denoted $\mathcal{F}$), is commonly used. Each subset of variables in the FOS corresponds to a set of linked variables that are jointly optimized by using the subset as a crossover mask during variation.
In GP-GOMEA, this FOS is learned during optimization based on a pairwise similarity measure between decision variables, such as MI. After estimating the similarity measure, hierarchical clustering with UPGMA~\cite{gronauOptimalImplementationsUPGMA2007} is performed to repeatedly merge subsets of variables starting from all single variable subsets. 
The last merge, i.e., the subset containing all variables, is discarded as the corresponding crossover mask would perform
replacement instead of variation. This hierarchical FOS structure is commonly referred to as a Linkage Tree (LT)~\cite{thierensLinkageTreeGenetic2010,dushatskiyParameterlessGenepoolOptimal2021a}.

Variation is performed separately for each solution, each of which is first cloned into an offspring solution. Next, for each offspring solution, each subset in the FOS is considered, and the offspring solution inherits all variables in the subset from a donor solution chosen uniformly randomly from the population. Specific to the GP domain, an evaluation is only necessary if this leads to a modification of any actively used part of the expression. If this is the case, the fitness of the resulting solution is compared with a backup and regressions in fitness are reverted.

\begin{algorithm}
\DontPrintSemicolon

\SetKwFunction{InitializeAndEvaluatePopulation}{InitializeAndEvaluatePopulation}
\SetKwFunction{ShouldTerminate}{TerminationCriteriaSatisfied}
\SetKwFunction{EstimateSimilarity}{EstimateSimilarity}
\SetKwFunction{LearnLinkageModel}{BuildLinkageTree}
\SetKwFunction{Shuffle}{Shuffle}
\SetKwFunction{Random}{Random}
\SetKwFunction{Clone}{Clone}
\SetKwFunction{Evaluate}{Evaluate}

\small
$\mathcal{P} \longleftarrow$ \InitializeAndEvaluatePopulation{}\;
\While{$\lnot$\ShouldTerminate}{
    \tcp{1. Linkage learning}
    $\mathcal{S} \longleftarrow$ \EstimateSimilarity{$\mathcal{P}$}\;
    $\mathcal{F} \longleftarrow$ \LearnLinkageModel{$\mathcal{S}$}\;
    \tcp{2. GOM}
    $\mathcal{O} \longleftarrow \Clone(\mathcal{P})$\;
    \For{$i \in \{1,\dots,|\mathcal{O}|\}$}{
            \For{$s \in$ \Shuffle{$\mathcal{F}$}}{
                $donor \longleftarrow$ \Random{$\{1,\dots,|\mathcal{P}|\} \setminus \{ i \}$}\;
                $backup \longleftarrow \Clone(\mathcal{O}_i)$\;
                $\mathcal{O}_i[s] \longleftarrow \mathcal{P}_{donor}[s]$\;
                \If{$\mathcal{O}_i[active] \neq backup[active]$}{
                    \Evaluate{$\mathcal{O}_i$}\;
                \If{$f(\mathcal{O}_i)$ worse than $f(backup)$}{
                    $\mathcal{O}_i \longleftarrow backup$
                }
                }
            }
        }
    $\mathcal{P} \longleftarrow \mathcal{O}$\;
}
\caption{\textbf{GP-GOMEA}\label{code:gpgomea}}
\end{algorithm}

To avoid determining an appropriate population size for a problem, (GP-)GOMEA is often paired with the interleaved multistart scheme (IMS), in which populations of exponentially increasing sizes perform generations in a scheduled manner, with larger populations performing generations less frequently than smaller populations, and under-performing populations are terminated when converged or once larger populations perform better~\cite{virgolinImprovingModelbasedGenetic2021a}.

\subsection{Linkage Measures in GP-GOMEA}

The FOS that dictates how variation in GP-GOMEA adheres to the learned variable dependencies, is constructed using a pairwise similarity measure. This measure is computed by interpreting the population as a dataset without regard for the mapping to the tree template. In this work, two new approaches are proposed to compute more fitting similarity measures by exploiting so far unused domain knowledge: the knowledge about inactive variables and the fixed template structure.

\subsubsection{Existing measures}

For GP, the previously explored measures are MI, an adjusted version thereof, and random noise~\cite{virgolinScalableGeneticProgramming2017,virgolinImprovingModelbasedGenetic2021a}. For the random measure, pairwise similarity is sampled from \(\mathcal{U}(0,1)\). For two random variables X and Y, the MI is defined as:
\begin{align}
MI(X, Y)= H(X) + H(Y) - H(X, Y)
\end{align}
where \(H(X) = \sum_{x}-P(X=x)\log_2(P(X=x))\) defines entropy.

In GP, expressions are typically initialized with domain-specific methods such as Half-and-Half~\cite{virgolinImprovingModelbasedGenetic2021a} instead of uniformly randomly. While this often achieves more desirable diversity in some aspects, such as the initial expression depths, biases can appear in the initial MI estimates~\cite{virgolinImprovingModelbasedGenetic2021a}. To avoid this, normalization factors were proposed to adjust the MI such that no linkage is detected directly after initialization, leading to increased performance~\cite{virgolinImprovingModelbasedGenetic2021a}.

\subsubsection{Exploiting knowledge about introns}
\begin{figure}[tbp]
    \centering
    \includegraphics[width=0.9\linewidth]{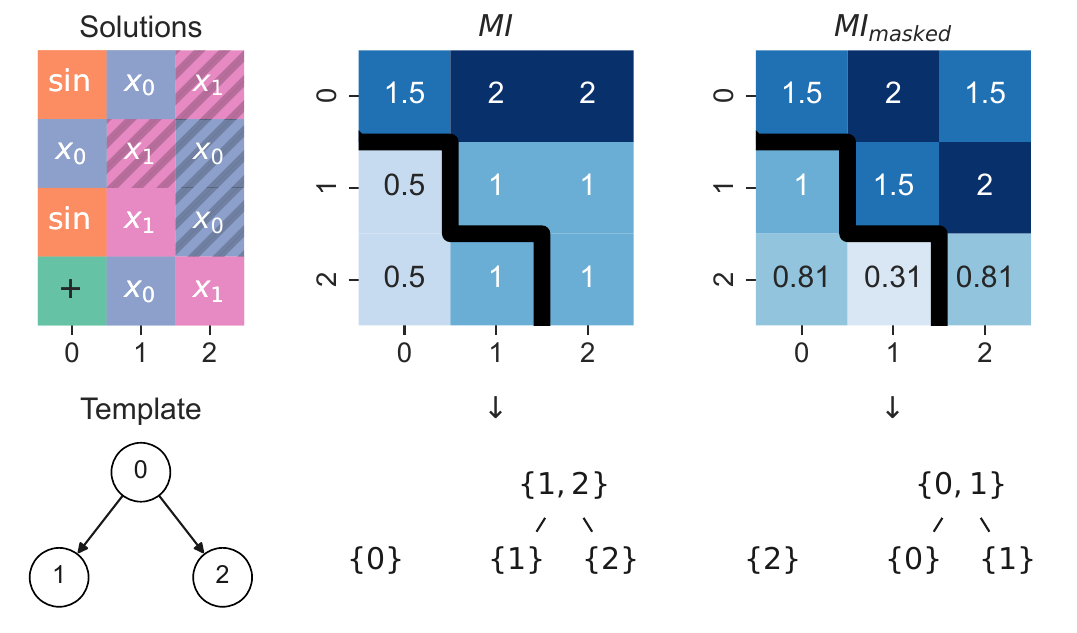}
    \Description{An example population of four solutions for a template with depth 1 (3 Nodes), together with the pairwise similarities for the proposed variant where introns are masked and normal mutual information. The example population contains inactive variables, leading to different similarities and thus different linkage trees estimated with the two similarity measures.}
    \vspace*{-4mm}
    \caption{Top left: population of 4 solutions. Bottom left: tree template. Matrices in blue show entropy (above the black line) and MI (below the black line). The corresponding linkage tree for both the normal and proposed masked MI are shown as well. The masked version treats all introns (shaded) as the same separate "masked" symbol.}
    \label{fig:linkage}
    \vspace{-0.1cm}
\end{figure}

The use of MI in GOMEA can often lead to successfully detecting variable dependencies. However, inactive variables in the form of introns are inherent to GP-GOMEA. The impact that this may have on dependency detection, has so far not been considered. For each solution, it is known which variables are currently active or inactive. The values of inactive variables do not influence the semantics and thus are freely interchangeable and not subject to selection pressure. From the perspective of entropy estimation, the values of inactive variables can therefore be considered to be noise, hiding the true signal of active variables.
To avoid this undesirable effect, we propose to mask the inactive variables using a special intron label during the entropy calculation, i.e., the alphabet of any variable now includes the "masked" label. An example is shown in~\Cref{fig:linkage}, where masking introns leads to a different linkage tree compared to using normal MI.
Masking makes the estimated MI more accurate, by ensuring that the estimate is fully based on active variables under selection pressure.
Note that this approach is not compatible with the adjustments for biased initialization methods proposed in~\cite{virgolinImprovingModelbasedGenetic2021a}. This is because initially, variables that are inactive in the entire population would lead to divisions by zero.

\subsubsection{Exploiting the known template}

In discrete and real-valued evolutionary optimization literature, it is common
to model dependencies with a variable interaction graph
(VIG)~\cite{whitleyGrayBoxOptimization2016b,tinosIteratedLocalSearch2024,andreadisFitnessbasedLinkageLearning2024}. In such a graph,
each variable is a vertex, and an edge exists between two vertices if there
exists a dependency between the two variables. Commonly, a VIG is associated
with grey-box optimization scenarios in which it is assumed that the problem
being optimized is sufficiently known. By studying the problem formulation,
dependencies between variables may then be directly gauged by identifying which
variables are jointly part of a subfunction. As an extension, or more informed,
variant of the VIG, the edges in the graph can be weighted. The resulting wVIG
then expresses not only whether or not two variables are jointly dependent, but
also how strong that dependence is, relative to other dependencies. In GOMEA,
such a wVIG can be used directly to build a linkage tree (or other types of
linkage structures such as conditional linkage
sets~\cite{andreadisFitnessbasedLinkageLearning2024} by
creating a similarity matrix $\vec{S}$ that contains the weight of an edge between
vertices $i$ and $j$ at $\vec{S}_{ij}$ and $\vec{S}_{ji}$.

The function represented by a GP tree template in GP-GOMEA, is a nested function. It can be defined
using two functions $f$ and $g$ in which the first argument of $f$ determines the
operation represented in the corresponding node of the GP tree, which may
include returning the value of a constant or input variable and ignoring the other arguments. Function $g$ is used
only in leaf nodes of the GP tree template, returning the value of a constant or an input variable.
Given a data vector $\vec{d}$, the function $GP(\vec{x})$ that a tree template
represents, can be written as a nested application of function $f$ with function
$g$ at the deepest level. I.e., for depth 2, we have:
\begin{align}
  \!\!GP(\vec{x},\!\vec{d}) \!=\! f\!(x_0,\!f\!(x_1,\!g(x_3,\!\vec{d}),\!g(x_4,\!\vec{d}),\!\vec{d}),\!f\!(x_2,\!g(x_5,\!\vec{d}),\!g(x_6,\!\vec{d}),\!\vec{d})\!)\label{eq:subfunctions}
\end{align}
This function is the core of what imposes dependencies between the variables that
we optimize over in GP-GOMEA. Consider the SSE optimization function for symbolic regression over all
data vectors $\vec{d}^{i}$ and targets $t^i$:
\begin{align}
  \min_{\vec{x}}\left\{\sum_i \left(GP(\vec{x},\vec{d}^i)-t^i\right)^2\right\}
\end{align}
Essentially this is a large sum of independent terms. Focusing on a single term,
we find $GP(\vec{x},\vec{d}^i)^2 - 2GP(\vec{x},\vec{d}^i)t^i + (t^i)^2$.
Since the variables in $GP$ are already
fully coupled through nonlinear interactions due to the nested function composition,
squaring the function does not alter the dependency structure, but amplifies existing
interactions. Hence, we can construct the wVIG based on the definition of $GP(\vec{x},\vec{d})$.

\begin{figure}[tbp]
    \centering
    \includegraphics[width=\linewidth]{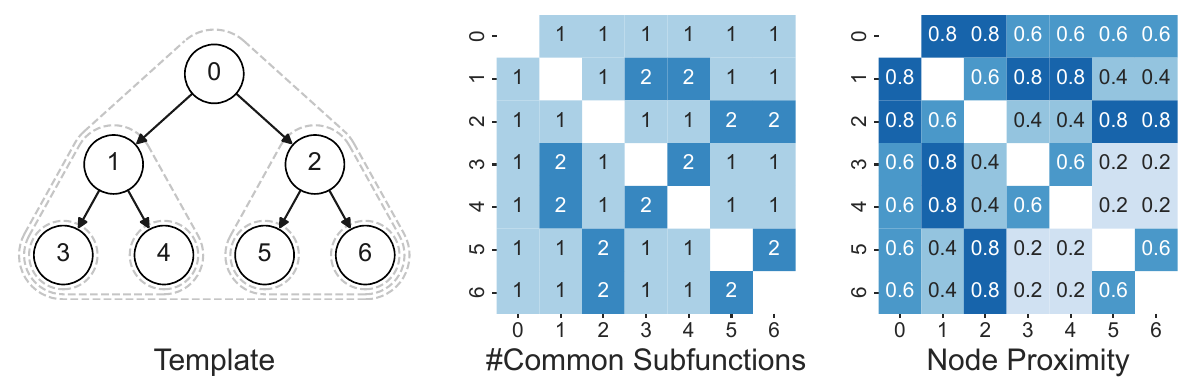}
    \Description{An example template for a full binary tree with depth 2, where subfunctions are indicated and nodes are enumerated breadth-first. Each leaf (nodes 3,4,5,6) is a subfunction, the nodes at depth 1 are larger subfunctions (node 1 contains 3,4 and node 2 contains 5,6) and the root (node 0) subfunction contains all other nodes. The figure also shows the subfunction based measure, where the pairwise similarity is the number of common subfunctions. Finally, the figure also shows the node proximity measure. For example, the node distance between the root (position 0) and the right child (position 2) is 1 edge. The maximum distance is four edges, for example from the left-most and right-most leaf (position 3 and 6 respectively) have the root as the common ancestor, leading to a total of four edges between the two positions (2 times 2 edges to the root). The example the node proximities obtained after normalizing by dividing by 1 + the maximum distance and inverting the distance.}
    \vspace*{-7mm}
    \caption{An example template with highlighted subfunctions and the corresponding node proximity and subfunction based similarity measures.}
    \vspace{-0.3cm}
    \label{fig:node_proximity}
\end{figure}

A straightforward and common way of doing so is counting how many times a variable
is contained in the same subfunction as another variable. For the depth 2 case, we get the similarity matrix defined in the middle of Figure~\ref{fig:node_proximity}.

However, for multiple reasons, the dependence between the arguments of a parent node in the template tree is
likely almost always smaller than the dependence between an argument and its
parent node. One of these reasons is that some functions may impose no dependencies between its
arguments. Another reason is that generally, the higher a node is in the template
tree (and therefore the less deeply nested in the functional description), the
larger its impact on changing the output of the tree and thus on the
regression error. Moreover, internal nodes may be terminals. Thus,
the second and third arguments of $f$ may not always be used. In that case, it does not matter what
values the variables take in the deeper nested functions (as they are then introns),
and thus, in those cases, those variables are independent. We can convey this information in a
systematic manner by defining a distance function between any two nodes in the
GP template that is the smallest distance between these nodes to travel along the template. This
makes the distance between two arguments of the same parent node 1 larger than
the distance between each of the arguments and the parent node (i.e., 2 and 1, respectively).

This distance measure can be turned into a normalized similarity measure as follows:
\begin{align}
  \vec{S}_{i,j} =
  \begin{cases}
    1 - \frac{d(i, j)}{1 + \max_{k,l \in \mathcal{I}}\{d(k,l)\}} & \text{if }i\text{ and }j\text{ are connected}\\
    0 & \text{otherwise}
  \end{cases}
\end{align}
where the distance \(d(i,j)\) is the total number of edges between nodes \(i\) and \(j\) and their closest ancestor in the template, and \(\mathcal{I}\) is the set of all node indices.

As shown on the right of~\Cref{fig:node_proximity}, while nodes are still equally similar by definition
using this measure of structural relatedness, there is more granularity than in the direct
subfunction encapsulation approach. 
During the hierarchical clustering,
ties are broken randomly if there are multiple merges with the same simiarlity. By rebuilding the linkage tree every generation, different LT FOS structures representing the similarity matrix are created. 
Note that compared to MI estimates, this form of pre-defined similarity disregards semantic relations due to variable values entirely. However, there is no need for population sizes sufficiently large to get statistically reliable estimates and literature has shown that appropriate offline similarity measures can lead to much better results~\cite{tinosGeneticAlgorithmLinkage2023,bouterPitfallsPotentialsAdding2025,bouterLeveragingConditionalLinkage2020}.

\section{Experiments and Results\label{sec:experiments}}

In this section, two experiments are performed to evaluate the proposed measures. First, the proposed measures are compared with previously used measures. The second experiment aims to uncover what linkage structures are learned.

\subsection{Experimental Setup}

To isolate the effect of linkage learning from other algorithm settings such as the fixed structural template and whether linear scaling~\cite{keijzerScaledSymbolicRegression2004} (hereafter LS) is enabled, all combinations are tested. LS is a much‑used post‑processing technique that provides scale and translation invariance by performing a linear regression on the output of the GP expression. The other settings are detailed in~\Cref{tab:setup}. The operator set was chosen based on~\cite{nicolauChoosingFunctionSets2021a}. The coefficient of determination (\(R^2\) Score) is used as objective:
\begin{align}
   R^2(y_{pred},y_{target}) &= 1 - \frac{MSE(y_{pred}, y_{target})}{var(y_{target})}\\[-0.75ex]
   MSE(y_{pred}, y_{target}) &= \frac{1}{N}\sum_{i=1}^N(y_{pred_i} - y_{target_i})^2
\end{align}
where \(y_{pred}\) and \(y_{target}\) are the predicted output (as in~\Cref{eq:subfunctions}) and target variable respectively, and \(N\) is the dataset size. Note that with LS enabled, the predicted values are scaled and transformed before computing the prediction error.

The problems we used in our experiments (listed in~\Cref{tab:datasets}) were selected to contain varying numbers of input features.
For each problem, 25\% of the instances form a held‑out test set. The remaining data is split into 5 cross‑validation folds, using 80\% for training and 20\% for validation. Each fold is repeated 6 times with different seeds, yielding 30 independent runs. Seeds are shared across algorithms so that all algorithms have identical initial populations.

\begin{table}[tb]
  \caption{The problems used in the experiments.}
  \vspace*{-3mm}
  \label{tab:datasets}
  \scalebox{0.9}
  {
  \begin{tabular}{lcc}
    \toprule
    Dataset & \#Instances & \#Features \\
    \midrule
    Airfoil~\cite{thomasbrooksAirfoilSelfNoise1989a} & 1503 & 5  \\
    Bike Sharing (Daily)~\cite{fanaee-tBikeSharing2013} & 731 & 12 \\
    Concrete Compressive Strength~\cite{i-chengyehConcreteCompressiveStrength1998a} & 1030 & 8 \\
    Dow Chemical~\cite{SymbolicRegressionCompetition2012} & 1066 & 57 \\
    Tower~\cite{vladislavlevaOrderNonlinearityComplexity2009}
    & 4999 & 25 \\
    \bottomrule
  \end{tabular}
  }
\end{table}

Numerical constants in the form of ephemeral random constants (ERCs) are used for which random values are sampled during initialization that remain constant throughout evolution.
During linkage learning, ERCs are handled as in~\cite{schlenderImprovingEfficiencyGPGOMEA2024}, i.e., for the MI-based measures, constant values are split into 25 
discrete bins (note: measures not considering population statistics are unaffected).

In the first experiment, a computational budget of \(10^7\) evaluations was used. As in~\cite{schlenderImprovingEfficiencyGPGOMEA2024}, the IMS starting population size is set to 64 with a subgeneration factor of 10 generations before larger populations are started.
All experiments follow the setup described in~\Cref{tab:setup} and were run on a separate physical core of a machine with two Intel Xeon E5-2699v4 processors, ensuring that runtimes are comparable.

In the second experiment, instead of IMS, a single population with a fixed size of 1024 and a generational budget of 20, a template height of 5 (31 Nodes), and LS enabled, is used. This population size aligns with earlier work on GP-GOMEA~\cite{virgolinImprovingModelbasedGenetic2021a,virgolinCoefficientMutationGenepool2022,kochSimultaneousModelBasedEvolution2024a}. 
\begin{table}[bp]
  \vspace*{-2mm}
  \caption{Experiment Setup}
  \vspace*{-3mm}
  \label{tab:setup}
  \scalebox{0.9}{
  \begin{tabular}{l|cc}
    \toprule
    \textbf{Setting} & \textbf{Experiment I} & \textbf{Experiment II} \\
    \midrule
    Objective & \multicolumn{2}{c}{$R^2$ Score} \\
    Initialization & \multicolumn{2}{c}{Half-and-Half~\cite{virgolinImprovingModelbasedGenetic2021a}} \\
    ERC Initialization & \multicolumn{2}{c}{\(\mathcal{U}(\text{min}(y_{train}), \text{max}(y_{train}))\)}  \\
    Operator set & \multicolumn{2}{c}{\(\{+,-,\times,\div,\sin\}\)} \\
    Termination & \(10^7\) evaluations & \(20\) generations \\
    Linear scaling & Yes / No & Yes \\
    Template Height & 5 / 7 (31 / 127 Nodes) & 5 (31 Nodes)\\ 
    \bottomrule
  \end{tabular}}
  \vspace{-0.3cm}
\end{table}

We consider the following linkage measures: random linkage, MI, bias adjusted MI as in~\cite{virgolinImprovingModelbasedGenetic2021a}, the proposed approach of masking inactive variables when computing the MI, and the similarity based on node proximity in the structural template with and without repeated building of the linkage tree every generation, hereafter abbreviated as Random, MI, $MI_{adjusted}$, $MI_{masked}$, Node, and Node (static) respectively. Both building of the linkage tree only once and rebuilding it every generation are considered for the node proximity measure to assess the effect of the aforementioned random tie-breaking during hierarchical clustering. Results for another baseline assuming full variable independence and the subfunction based measure are presented and discussed in the supplementary material, where the results confirm that variables are not independent and the Node measure outperforms the subfunction based measure.

To analyze the results, we follow the recommendations and implementation provided by~\cite{agarwalDeepReinforcementLearning}, where interval estimates via stratified bootstrap confidence intervals are preferred over point estimates and statistical significance tests to take uncertainty and the effect size into account. We mainly consider training performance, as that is directly optimized and we are primarily interested in the effectiveness of the linkage models instead of the generalization performance. Statistical analysis is also done as recommended in~\cite{demsarStatisticalComparisonsClassifiers2006} with the implementation provided by~\cite{herboldAutorankPythonPackage2020}.

\subsection{Experiment 1: Effectiveness of Various Linkage Measures}
\begin{figure}[bp]
    \centering
    \includegraphics[width=\linewidth]{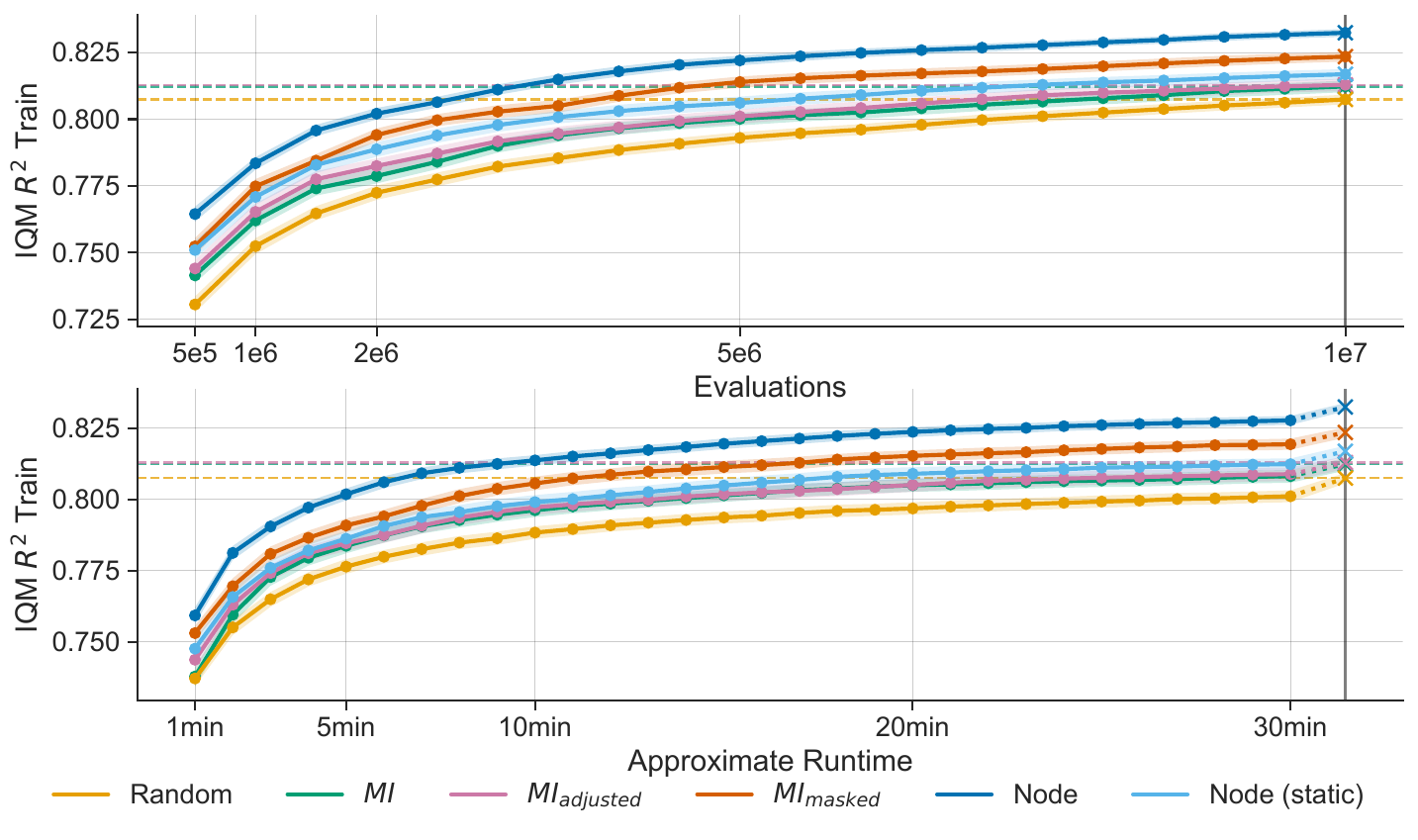}
    \Description{Convergence plots are shown using the aggregate interquartile mean (IQM) measure across all problems and settings considered. For both the number of evaluations and the approximate runtime, several steps are shown where the relative ordering of the IQM does not change. Node performs best, followed by $MI_{masked}$, Node (static), $MI_{adjusted}$, MI and finally Random performs worst. Node reaches the final aggregate IQM accuracy obtained by the existing methods already after approximately 3.5 million evaluations/10 minutes, which corresponds to roughly one third of the total evaluation budget/time taken respectively.}
    \vspace{-0.75cm}
    \caption{The interquartile mean (IQM) training $R^2$ and bootstrapped 95\% confidence intervals (as per~\cite{agarwalDeepReinforcementLearning}) over evaluations and approximate runtime, across all runs performed. For the runtime, the last recorded value before each point in time was used, and dashed horizontal lines show the final values reached for the existing measures. The vertical line indicates the computational budget (not to scale w.r.t runtime) where the final values reached are marked with crosses.}
    \vspace{-0.25cm}
    \label{fig:convergence}
\end{figure}
\begin{figure*}[htbp]
    \centering
    \includegraphics[width=\linewidth]{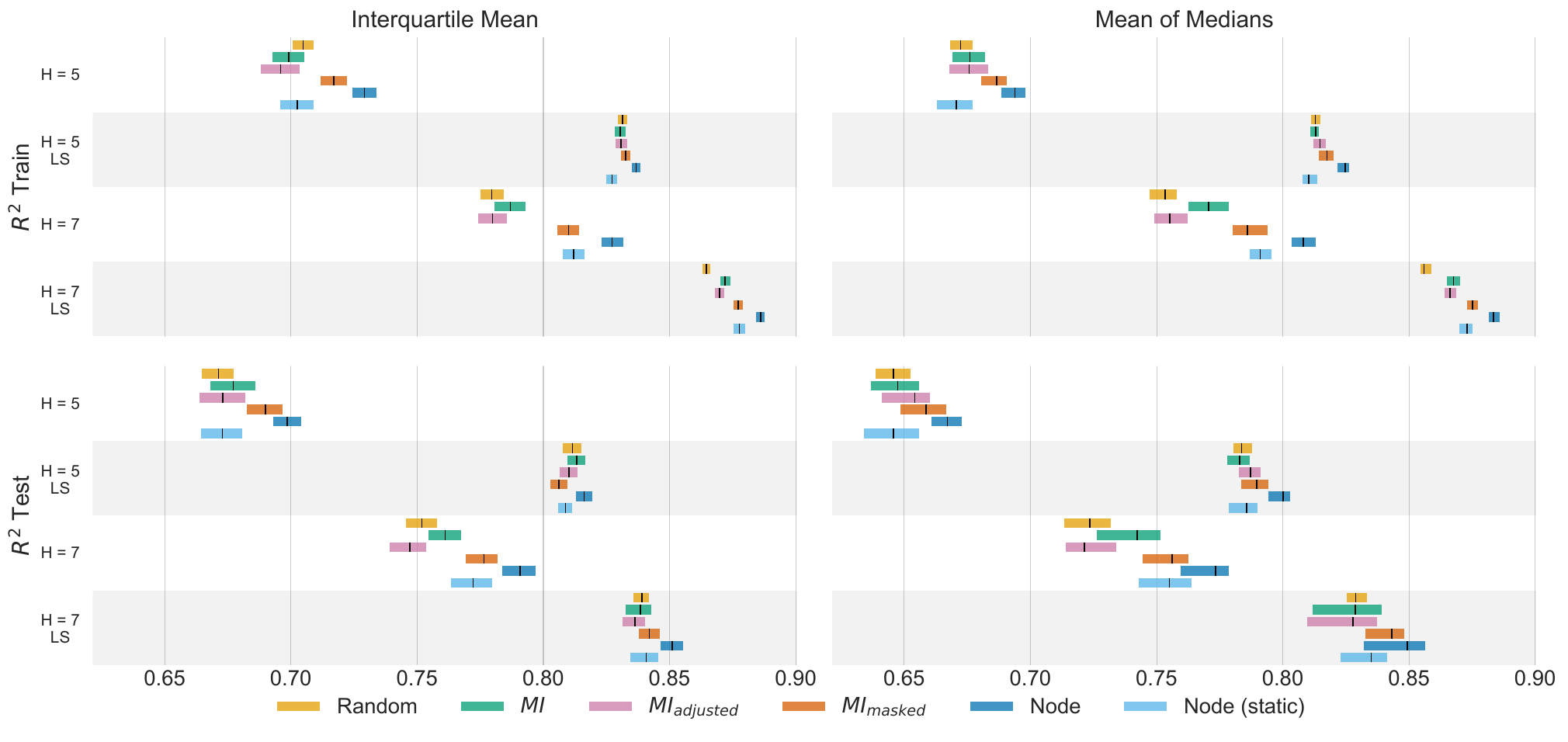}
    \Description{The figure shows the aggregate $R^2$ scores (higher is better) on the problems for each combination of template height and LS considered with confidence intervals. For each combination, both for the training and testing accuracy, there tends to be one group of less accurate measures containing Random, $MI_{adjusted}$ and MI (mostly in this order with improving accuracy). $MI_{masked}$ tends to perform clearly better, with Node generally outperforming all other methods. The static version of Node performs worse compared to node, sometimes better than the group of existing methods and sometimes comparably. With linear scaling enabled all methods are closer together, and the differences between methods increase with the larger templates using height 7.}
    \vspace{-0.85cm}
    \caption{Aggregate $R^2$ scores (higher is better) on the problems for each combination of template height and LS considered. The interquartile mean corresponds to the mean after discarding the bottom and top 25\% of runs for each problem, and the mean of medians corresponds to the mean of the median performances on each problem. The colored bar corresponds to the 95\% confidence interval estimated using a percentile bootstrap with stratified sampling as per~\cite{agarwalDeepReinforcementLearning} with the expected value in black.
    }
    \label{fig:interval_estimates}
\end{figure*}
\begin{figure*}[ht]
    \centering
    \vspace*{-2mm}
    \includegraphics[width=\linewidth]{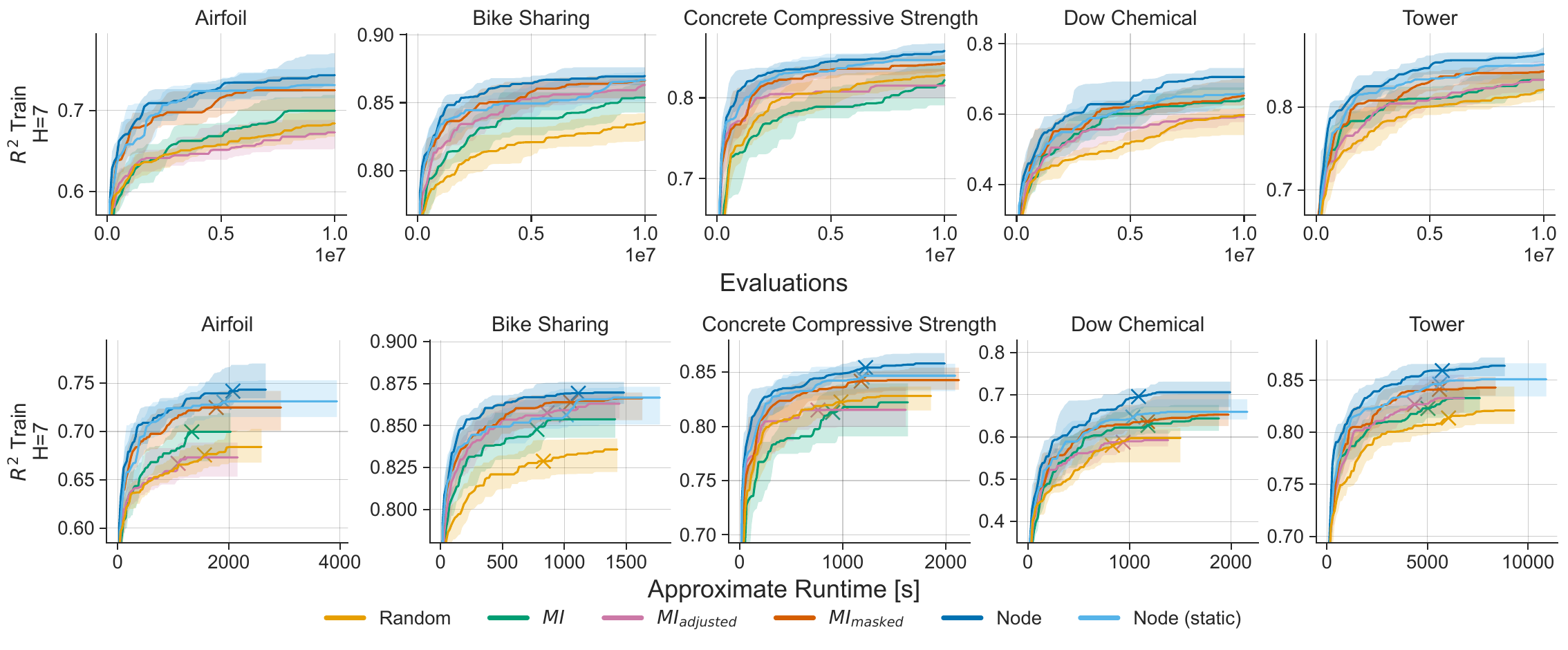}
    \Description{The figure shows median and interquartile range on all problems with template height 7 and without linear scaling. While there is variation between problems (on Concrete Compressive strength, MI performs worst and on dow chemical $MI_{masked}$ does not outperform the existing methods), the general order with decreasing final accuracy is Node performs best, followed by $MI_{masked}$, Node (static), $MI_{adjusted}$, MI and finally Random performs worst. All methods tend to have an approximately logarithmic gain in accuracy, i.e. initially fast and then exponentially more computational effort is needed for similar gains. The proposed methods reach the final accuracies of the existing methods considerably faster, with Random generally performing clearly worse than all other methods.}
    \vspace{-0.8cm}
    \caption{Median training $R^2$ score (higher is better) over evaluations and approximate runtime across problems with template height 7 and without linear scaling. The filled area corresponds to the interquartile range, and the runtime where the first run finished is marked with a cross, after which the final value obtained is re-used for completed runs until all runs have finished.}
    \label{fig:convergence_h7nols}
\end{figure*}

\begin{figure*}[htbp]
    \centering
    \includegraphics[width=0.875\linewidth]{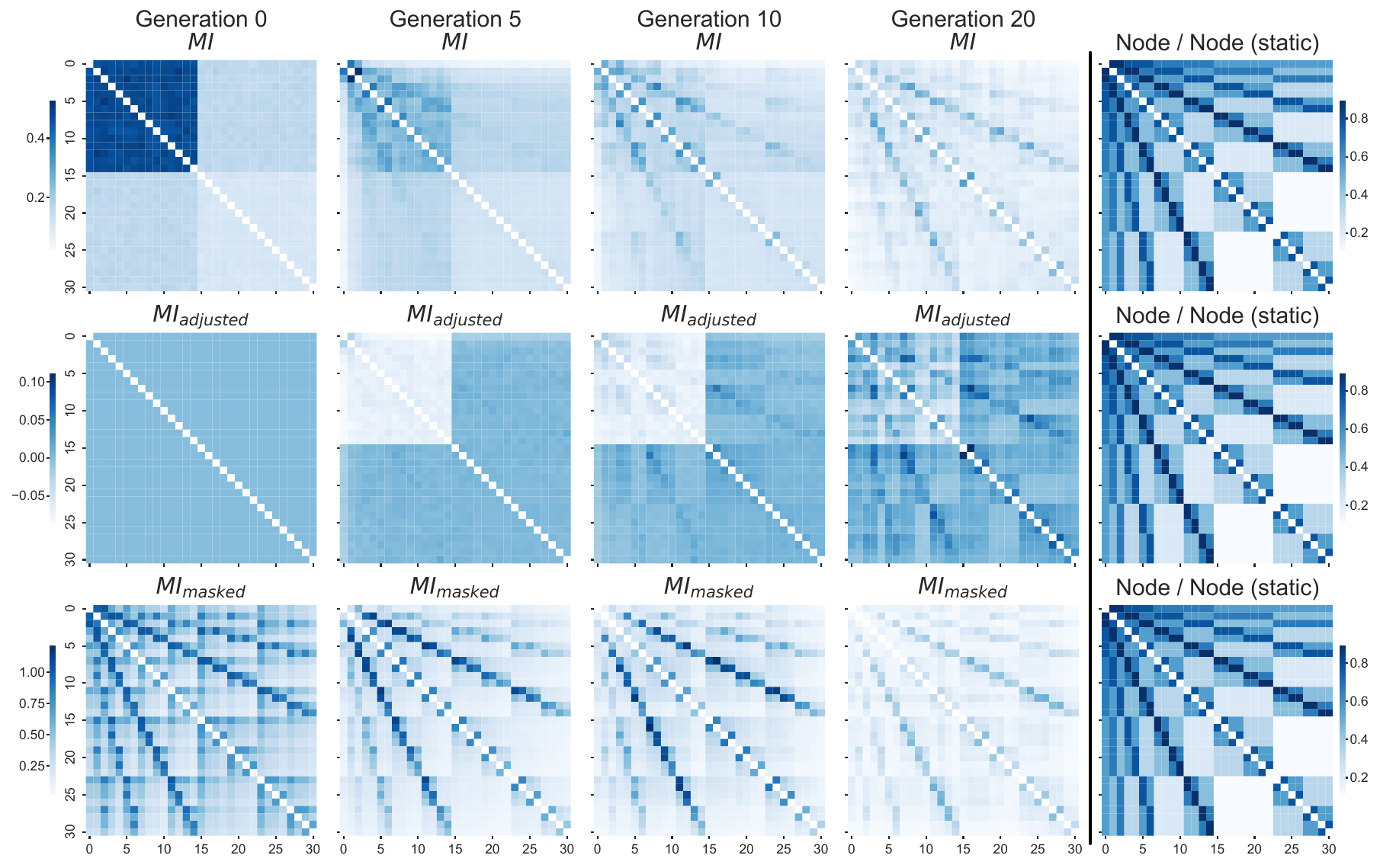}
    \Description{For generations 0 (i.e. after initialization and before variation), 5, 10 and 20, the averaged pairwise similarity measures are shown as heatmaps. Initially MI shows clear bias, while $MI_{adjusted}$ assumes no linkage between variables in the first generation. Both $MI_{masked}$ is very similar to Node, showing that the template structure is learned. With increasing generations, both MI and $MI_{adjusted}$ also increasingly resemble the Node measure, albeit less distinctively.}
    \vspace{-0.3cm}
    \caption{The different linkage measures over generations, averaged over 30 runs on the Bike Sharing dataset with a template height of 5 (31 Nodes), LS and a fixed population size of 1024. Note that different scales are used for each measure, as the focus lies on differences between variable pairs within measures instead of differences between linkage measures. The Node measure is shown only once per row as it does not change throughout evolution to ease comparisons.}
    \label{fig:similarity}
\end{figure*}
To assess the effectiveness of the various linkage measures, we consider both the final accuracies achieved (\Cref{fig:interval_estimates}) and intermediate results throughout optimization in~\Cref{fig:convergence,fig:convergence_h7nols}. The supplementary material provides per problem results (\Cref{appendix:problem_results}), and a repeat of the first experiment showing that the conclusions made extend to varying operator set sizes (\Cref{appendix:operators}).

The expected final $R^2$ scores on both the training and test splits are shown in~\Cref{fig:interval_estimates} with two different aggregate metrics, split up into the various combinations of template height and linear scaling that we considered. Two different aggregates, based on the mean performance without outliers, and the median per problem performance are provided, as both measures are differently affected by asymmetric accuracy distributions.

Across all settings, the Node measure performs best, followed by $MI_{masked}$, indicating that the proposed methods provide more effective linkage information. 
Interestingly, the static version of the Node measure, where the initial linkage tree is re-used within the same population for all subsequent generations, clearly performs worse than the randomized version. This aligns with~\cite{thierensPredeterminedLearnedLinkage2012}, where it was found that there is not necessarily one best static linkage structure and randomization can be beneficial. This could be due to the static version introducing bias by breaking ties only one way, while the repeated building of linkage trees with ties being broken randomly, leads to an unbiased FOS representation of the estimated linkage structure over multiple generations. Another effect is that the randomization possibly promotes propagation of building blocks between overlapping subsets of subsequent FOS structures better than the static version.

In general, increasing the template height and thus the possible expression sizes, leads to better accuracies, and the same holds for enabling LS. While the template size and LS have a larger impact on accuracy, clearly, linkage learning too has a noticeable impact. However, considering only the already existing methods, the random baseline is not clearly outperformed for all settings, suggesting that previously proposed linkage measures are suboptimal for GP.

When it comes to the relative performance of the linkage measures, increasing the number of decision variables (template height) widens the gap between the various methods. LS, as arguably could be expected, has the opposite effect and decreases the differences. Moreover, of course, there is a generalization gap when the training and test performance are compared, however, the relative performance of the various linkage measures mostly remains unaffected.
Overall, MI tends to outperform the bias-adjusted version, while the setting using a template height of 5 and LS as used in~\cite{virgolinImprovingModelbasedGenetic2021a} where statistical tests showed the opposite.
Compared to the experiments performed in~\cite{virgolinImprovingModelbasedGenetic2021a}, the budget used here is an order of magnitude larger, and thus any beneficial effects of mitigating initial biases are likely less visible in the final results achieved after exhausting the computational budget. \Cref{fig:convergence}, where the convergence aggregated over all combinations and problems considered is shown, also shows an initial advantage of $MI_{adjusted}$ over MI, aligning with the results obtained in~\cite{virgolinImprovingModelbasedGenetic2021a}. This suggests that the bias adjustment itself does work, however, as optimization progresses, the initial bias becomes less and continued mitigation can be harmful as the MI measure outperforms its adjusted counterpart in various settings. 

\begin{figure}[htbp]
    \centering
    \includegraphics[width=0.95\linewidth]{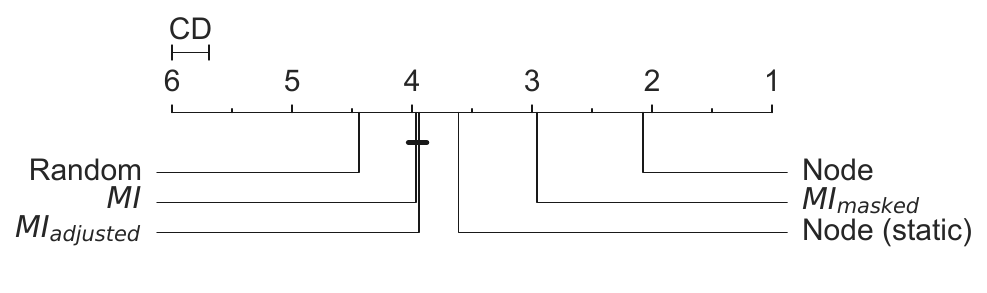}
    \vspace{-0.5cm}
    \Description{The average rank of methods is as follows from best to worst: Node, $MI_{masked}$, Node (static), $MI_{adjusted}$, MI, Random where all differences except MI vs $MI_{adjusted}$ are significant.}
    \caption{Statistical differences across all problem, template height and linear scaling combinations considered obtained by performing a Friedman test followed by the pairwise Nemenyi test using~\cite{herboldAutorankPythonPackage2020}. Lower ranks indicate better performance, and measures not connected by a bar exceed the critical difference (CD) and are significant (at \(p=0.05\)).}
    \label{fig:differences}
\end{figure}

Intermediate results and the runtime performance across all runs are shown in~\Cref{fig:convergence}, and \Cref{fig:convergence_h7nols} provides convergence results per problem for template height 7 without LS. Results for the other combinations are in~\Cref{appendix:problem_results} of the supplementary material. In both figures, the Node and $MI_{masked}$ measures outperform the other measures throughout evolution with respect to both evaluations performed and runtime. 
In terms of evaluations, use of the proposed measures outperforms use of the existing measures at only half the computational budget, and the same holds for comparing the IQM achieved using the Node measure after $5e5$ evaluations, and the existing methods after $1e6$ evaluations. Considering runtime, the varying number of instances between the problems (see~\Cref{tab:datasets}) leads to varying runtimes until the full budget is exhausted, ranging from several minutes to hours. Nonetheless, the Node and $MI_{masked}$ measures exceed the final accuracy reached by the existing methods after 10 and 20 minutes, respectively, showing a substantial speed-up in terms of runtime and highlighting the importance of exploiting linkage.
The observed efficiency improvement
possibly is proportional to the number of nodes in the template, as the advantage in~\Cref{fig:convergence_h7nols} is larger compared to~\Cref{fig:convergence} where all settings are aggregated.
Statistical differences are shown in~\Cref{fig:differences} across all settings considered, where a critical difference diagram is used to show the average measure ranks. It can be seen that all measures are significantly different from each other, except the normal and bias adjusted variants of mutual information. Clearly, the proposed linkage measures exploit the problem structure more effectively compared to the existing measures, with the template informed Node measure performing best followed by $MI_{masked}$.

\subsection{Experiment 2: What linkage structures are learned?}

The learned pairwise similarities pertaining to the various measures averaged over 30 runs on the Bike Sharing problem are shown in~\Cref{fig:similarity}. Results for the other problems are shown in~\Cref{appendix:similarity} of the supplementary material. Note that both Node versions use the same similarity measure that does not change throughout evolution and Random is not shown as it is not informative.
The similarities shown at generation 0 are recorded before any variation is performed, i.e. directly after initialization.

In the first row of~\Cref{fig:similarity}, the bias in the mutual information estimate due to non-uniform initialization (as addressed in~\cite{virgolinImprovingModelbasedGenetic2021a}) is clearly visible in the first generation, and notably absent for $MI_{adjusted}$. As the number of generations performed increases, the bias in MI (and the bias adjustment of $MI_{adjusted}$) becomes less visible and both measures learn similar patterns of linked variables.

Compared to the existing measures, $MI_{masked}$ does not seem to suffer from the same bias as MI, suggesting that not accounting for inactive variables is the cause of the bias. The linkage structure learned by $MI_{masked}$ is highly similar to the node proximity measure introduced. However, with increasing generations, parts of the template structure learned by $MI_{masked}$ degrade over time.
The patterns learned in later generations by MI and $MI_{adjusted}$, as well as the linkage structure learned by $MI_{masked}$ resemble the Node measure, suggesting that indeed the distance of variables in the fixed template is highly informative. Clearly, once introns are taken into account, this structure can be learned from population statistics.
\Cref{appendix:similarity} of the supplementary material shows that this general trend of mutual information based linkage learning approximating the template structure holds across all problems considered.

\section{Discussion\label{sec:discussion}}

Learning and exploitation of linkage information can be crucial for effective optimization.
We re-visited linkage learning in GP-GOMEA with the aim to more fully exploit the available problem knowledge, in particular knowledge about inactive variables and the user-defined structural template shared across all solutions. The experiments clearly show that the proposed measures outperform existing measures. This holds across all combinations of template heights and linear scaling considered, with clear speed-ups in terms of evaluations and runtime performance.

The proposed methods include no new algorithm settings, and either are a simple modification to how introns are handled during the calculation of the mutual information between variable pairs or require no runtime cost in the case of the node proximity measure. Notably, masking introns to improve the signal-to-noise ratio of variables not under selection pressure is not specific to GP, and can be applied to other settings where it is known which variables are inactive, as in~\cite{guijtExploringSearchSpace2024,przewozniczekHoplikeProblemNature2024,przewozniczekConditionalDirectEmpirical2025}.

For the first time in GP-GOMEA, we also explored what linkage structures are learned by the various measures. Interestingly, all mutual information based variants seem to approximate the structural template, suggesting that at least for the problems considered, the solution representation is more important than variable values when it comes to the dependency strengths between variables. However, there could be problems where this does not hold. %

Clearly, the best solutions differ between problems, suggesting that the importance of different parts in the template can vary as well. This could potentially still be captured by runtime measures, especially the masked measure that we proposed that takes into account the conditionally inactive variables that appear due to the use of a tree template. It would be interesting to see in future work if such hybrid linkage measures could improve performance further.

Another aspect to consider is the modeling of the learned linkage, i.e. why randomization of the node proximity performs better, and how much randomization of the FOS structure is best. Currently, the tie breaking in UPGMA only resorts to randomness in the face of equally strongly linked subsets of variables, which might not truly lead to FOS structures that over multiple generations form an unbiased representation of the linkage information.

Beyond improving the learning, modeling and exploitation of linkage, other factors potentially interacting with linkage learning in GP were not considered, but could be equally important. This includes, for example, representing numerical constants differently as done in~\cite{kochSimultaneousModelBasedEvolution2024a} and the analysis of possible effects on interpretability from an explainability viewpoint, which is not considered here. Nonetheless, the fixed template size used in GP-GOMEA typically restricts the search space to small, and thus more likely humanly interpretable expressions.

\section{Conclusion\label{sec:conclusion}}

Learning and exploiting linkage is a key mechanism of GP-GOMEA, an algorithm among the current state-of-the-art when it comes to SR. With this in mind, we revisited this core aspect of the algorithm, leading to two new linkage measures that exploit more of the available problem knowledge.

Our experiments, through which we evaluated the use of various linkage measures in the context of varying template sizes and the use of linear scaling, showed that by exploiting known problem information, clear accuracy and runtime performance improvements are possible across all combinations considered. Moreover, the linkage structures that are learned, align with the user-defined structural template available a priori, with the best performing linkage measure being the one that encapsulates exactly the structure of the template, requiring no additional on-line linkage learning from the population statistics. This establishes a clear bridge between GP-GOMEA and other linkage learning EAs, particularly GOMEA variants, highlighting the advantage of explicitly modeling compositional problem structure, a principle widely recognized in discrete and real-valued optimization.


\bibliographystyle{ACM-Reference-Format} 
\bibliography{main}

\appendix

\section{Additional Measures\label{appendix:other_measures}}

\begin{figure}[tbp]
    \centering
    \includegraphics[width=\linewidth]{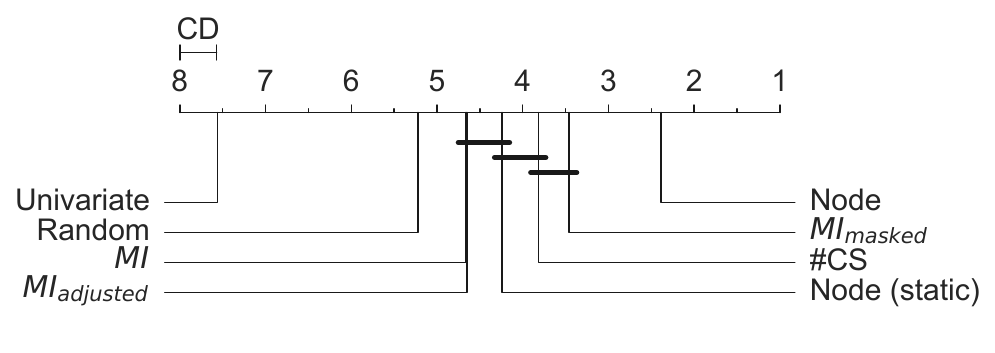}
    \vspace{-0.75cm}
    \Description{The average rank of methods is as follows from best to worst: Node, $MI_{masked}$, Common Subfunction Count, Node (static), $MI_{adjusted}$, MI, Random, Univariate where all differences except $MI_{masked}$ vs Common Subfunction Count, Common Subfunction Count vs Node (static), Node (static) vs MI/$MI_{adjusted}$, and MI vs $MI_{adjusted}$ are significant.}
    \caption{Statistical differences across all problem, template height and linear scaling combinations considered obtained by performing a Friedman test followed by the pairwise Nemenyi test using~\cite{herboldAutorankPythonPackage2020}. Lower ranks indicate better performance, and measures not connected by a bar exceed the critical difference (CD) and are significant (at \(p=0.05\)).}
    \label{fig:differences_all}
\end{figure}

In~\Cref{fig:differences_all}, and~\Cref{appendix:probability,appendix:problem_results,appendix:similarity,appendix:operators}, results are provided for all measures in the main work, and two additional measures. First, the subfunction based common subfunction count (as \#CS, with repeated LT rebuilding) and the univariate FOS (as Univariate), which assumes no variable interaction and consists of all single variable subsets. The univariate FOS is often used as baseline, and it can be the case that this FOS performs well even if there are variable interactions~\cite{bouterLeveragingConditionalLinkage2020}. However, thus far this FOS has not been considered in GP-GOMEA due to the fact that GP clearly is not a separable problem with independent decision variables. All results in the supplementary material include both additional measures, with the exeption for~\Cref{appendix:similarity}, where the univariate model is not shown as it assumes that there is no pairwise similarity. The results indicate that the univariate model clearly performs worse than all other methods indicating that linkage matters for GP. Furthermore, as expected, the Node measure which does not assume subfunction independence in the problem formulation generally outperforms the subfunction based measure.

\section{Probability of Improvement\label{appendix:probability}}

\begin{figure}[htbp]
    \centering
    \includegraphics[width=0.95\linewidth]{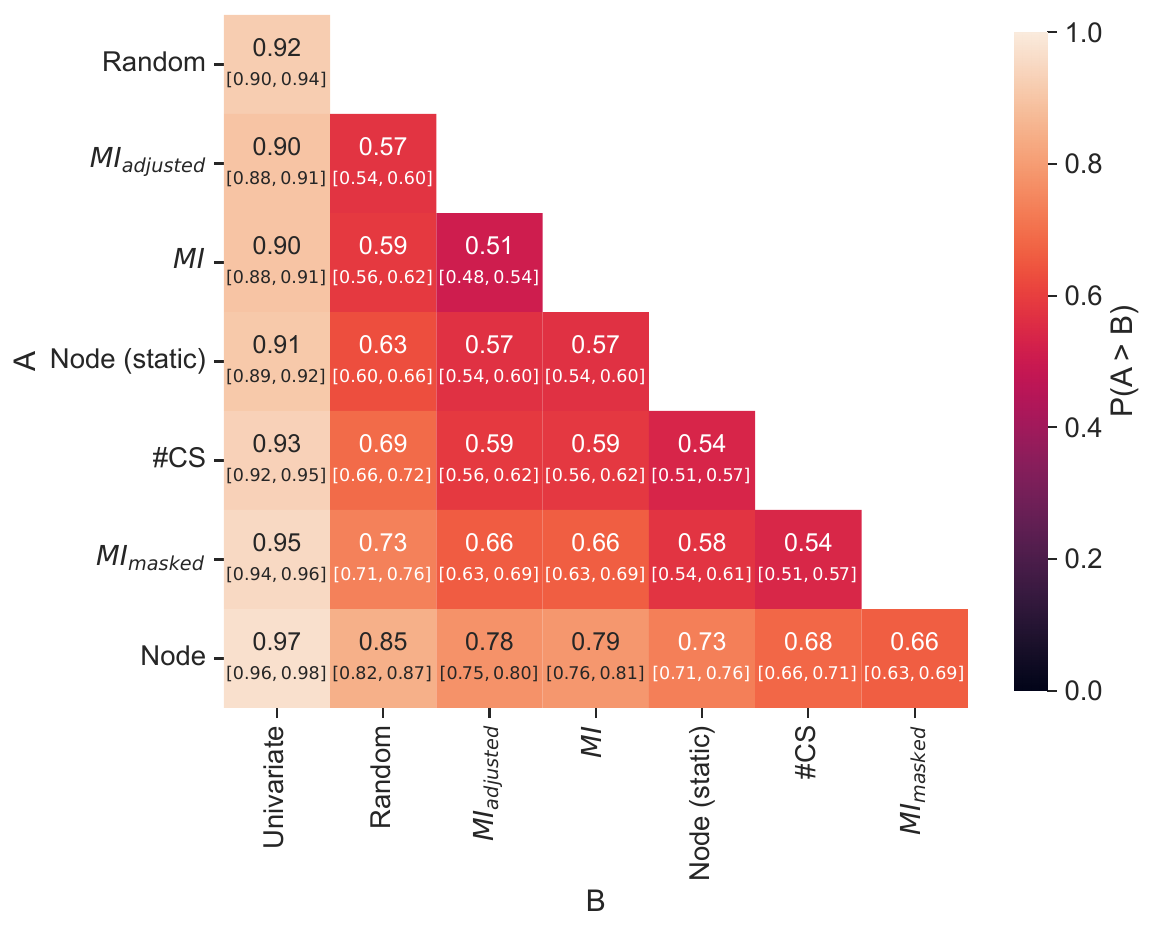}
    \Description{The figure shows the pairwise probabilities of one measure outperforming the other. Node has a roughly 80\% chance of performing better than all existing methods with narrow confidence intervals suggesting a worst case of a 75\% chance of outperforming the bias adjusted MI. Node has a 66\% chance of outperforming the static version and the intron aware masked version of MI, which in turn has a roughly 66\% chance of outperforming the existing methods again with narrow confidence intervals. Other than that, MI has a 51\% chance of outperforming $MI_{adjusted}$ with the confidence interval crossing the 50\% mark suggesting that the two measures are evenly matched. Both measures are roughly 60\% likely to outperform the random measure. All methods are roughly 90\% likely to outperform the Univariate FOS, and the common subfunction count performance is between Node (static) and $MI_{masked}$, i.e. better than existing methods, but not better than masking or Node.}
    \caption{The pairwise probabilities of measure A achieving a higher training $R^2$ score than measure B across all runs performed ($P(B \geq A) = 1 - P(A > B)$). The interval corresponds to the 95\% bootstrapped confidence interval as per~\cite{agarwalDeepReinforcementLearning}.}
    \label{fig:probabilities}
\end{figure}

In addition to the null-hypothesis based used in the main work, the pairwise probabilities of one measure performing better than another measure across all runs is shown in~\Cref{fig:probabilities}, computed using a stratified bootstrap using~\cite{agarwalDeepReinforcementLearning}. Probabilities with confidence intervals overlapping 50\% are clearly not significant and the results align with~\Cref{fig:differences}. In expectation, all measures outperform the univariate model clearly, as well as the random measure showcasing that linkage can be exploited in GP. Moreover, $MI$ and $MI_{adjusted}$ perform comparably with an edge towards $MI$. The static version of Node likely outperforms the existing measures, but not the other proposed methods. The subfunction based measure likely outperforms the existing measures, however, not the other measures proposed in this work except for the static version of the node proximity measure. $MI_{masked}$ is preferable to all measures except Node. Note that while Node is likely to perform best, there is overlap between the score distributions of all measures. For example, there is a roughly 15\% chance of Random outperforming Node after the full budget is exhausted.

\section{Per Problem Results\label{appendix:problem_results}}

\begin{figure*}[t]
    \centering
    \includegraphics[width=\linewidth]{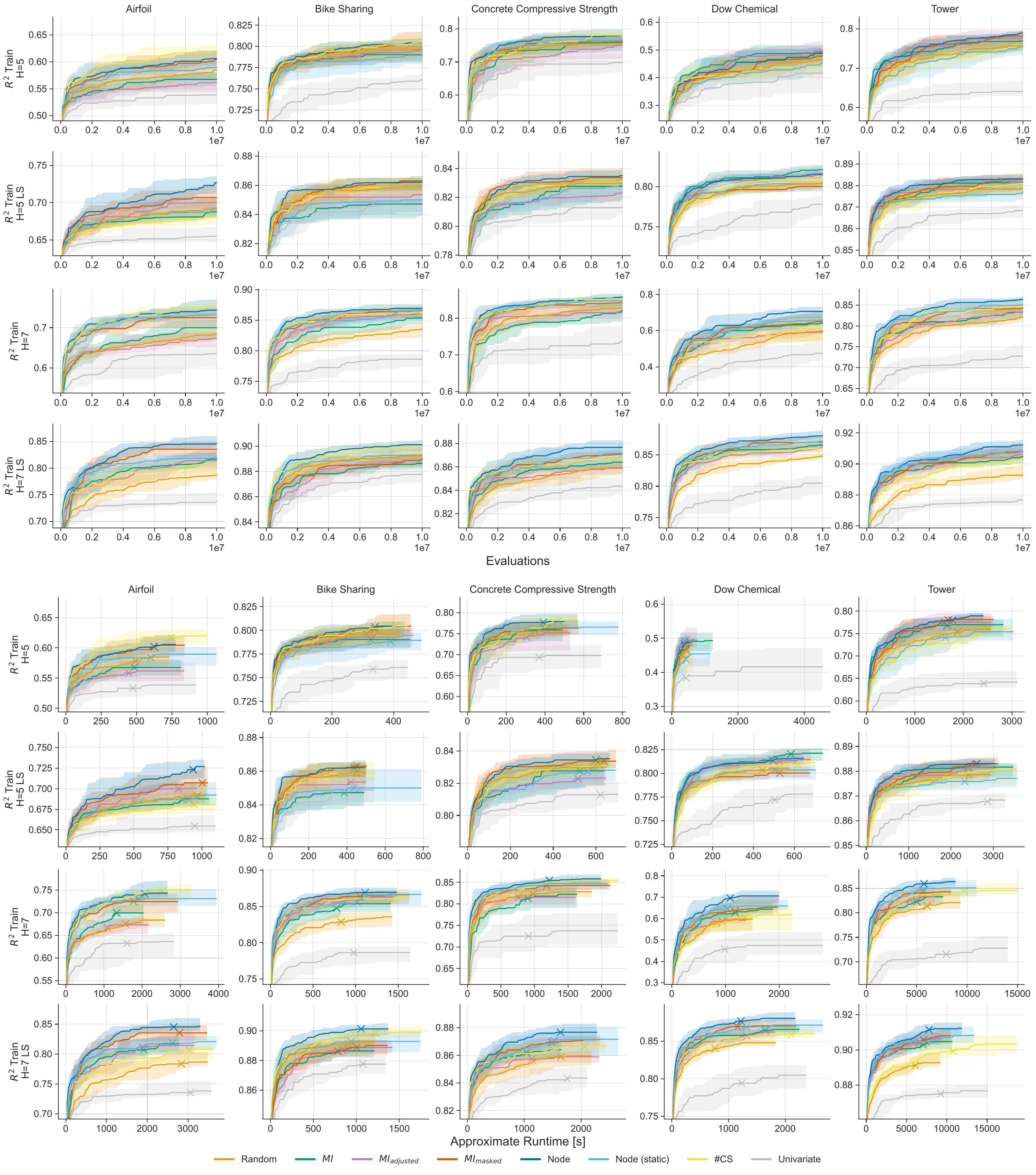}
    \Description{Convergence graphs for all problems and settings considered, showcasing the performance over time and evaluations. Generally, Node performs best, followed by $MI_{masked}$, Common Subfunction Count, Node (static), $MI_{adjusted}$/MI, Random and Univariate in roughly that order across problems.}
    \caption{Median training $R^2$ score (higher is better) over evaluations and approximate runtime across problem, template and linear scaling combinations. The filled area corresponds to the interquartile range, and the runtime where the first run finished is marked with a cross, after which the final value obtained is re-used for completed runs until all runs have finished.}
    \label{fig:per_problem_convergence}
\end{figure*}

\begin{figure*}[t]
    \centering
    \includegraphics[width=\linewidth]{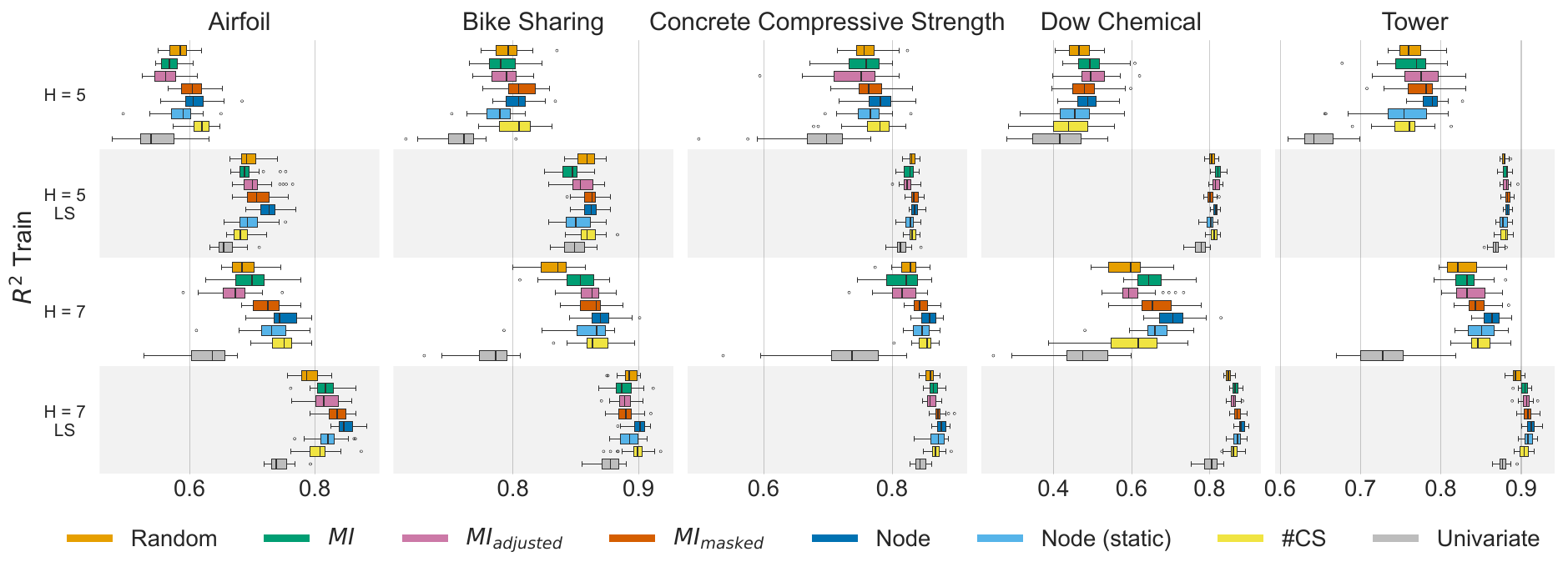}
    \vspace{-0.7cm}
    \Description{Final results for all problems and settings considered, showcasing the training accuracy. Generally, Node performs best, followed by $MI_{masked}$, Common Subfunction Count, Node (static), $MI_{adjusted}$/MI, Random and Univariate in roughly that order across problems.}
    \caption{$R^2$ score (higher is better) across problem, template and linear scaling combination considered. The boxes show the quartiles and the whiskers extend to points that lie within 1.5 inter-quartile ranges of the lower and upper quartile. Observations outside this range are displayed independently.}
    \label{fig:per_problem_scores}
\end{figure*}

\Cref{fig:per_problem_convergence} shows the training accuracy throughout evolution and \Cref{fig:per_problem_scores} shows the distribution of the final training $R^2$ scores achieved per problem. Notably, the proposed methods perform better in all settings except the Dow Chemical problem with a template height of 5. This problem has the largest number of input features, however, this likely it not the cause of the proposed methods underperforming as this is not the case with larger templates. Interestingly, while the Node measure typically reaches higher accuracies faster, the runs do not exhaust the evaluation faster. This possibly indicates that more GOM steps only modify inactive variables and thus no evaluation is performed. This effect is particularly evident for the Univariate model on the Dow Chemical problem without LS and template height 5, where some runs have a vastly longer runtime.

\section{Per Problem Similarity Measures\label{appendix:similarity}}

\begin{figure*}[htbp]
    \centering
    \begin{subfigure}[t]{0.495\linewidth}
        \centering
        \includegraphics[width=1.0\linewidth]{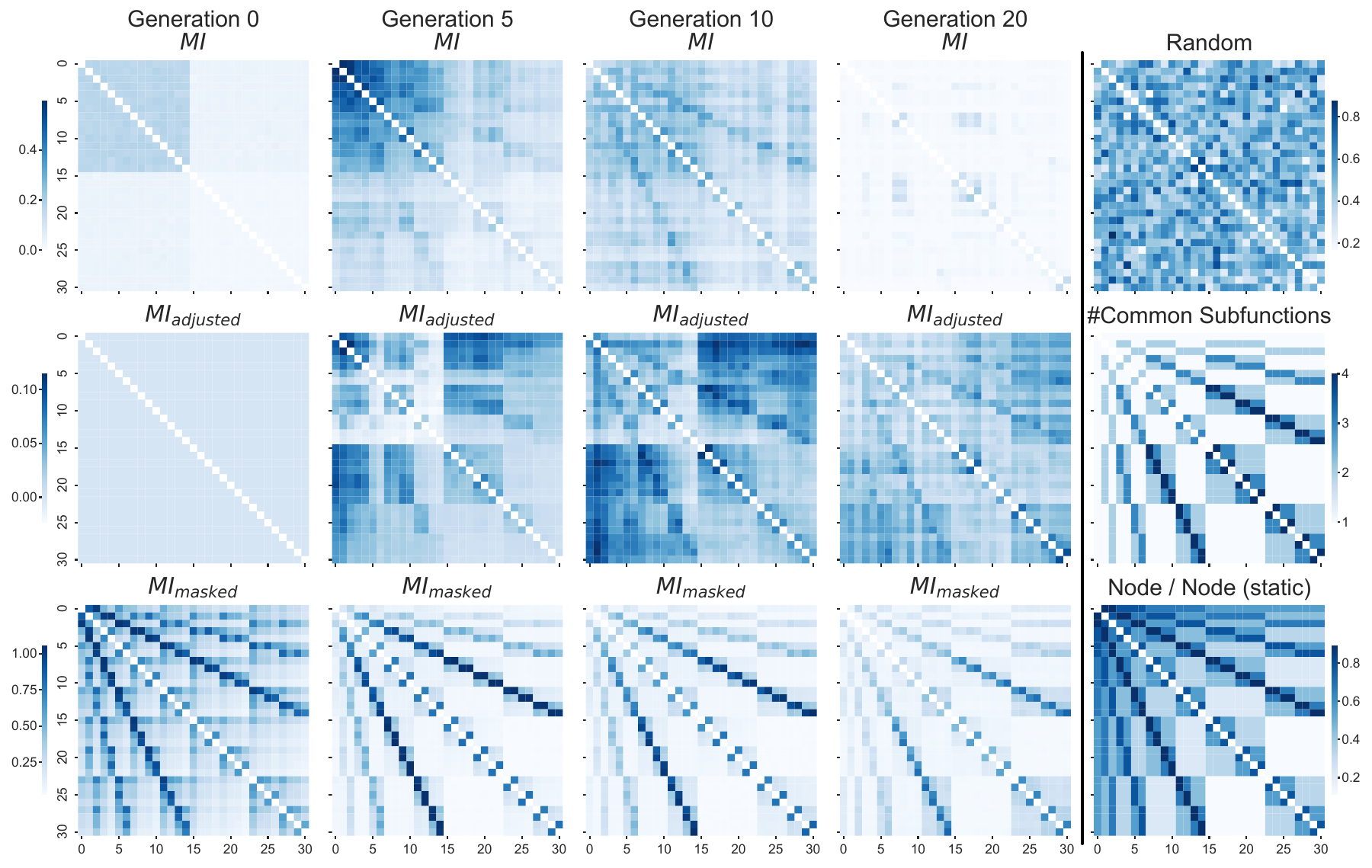}
        \caption{Airfoil}
    \end{subfigure}
    \hfill
    \begin{subfigure}[t]{0.495\linewidth}
        \centering
        \includegraphics[width=1.0\linewidth]{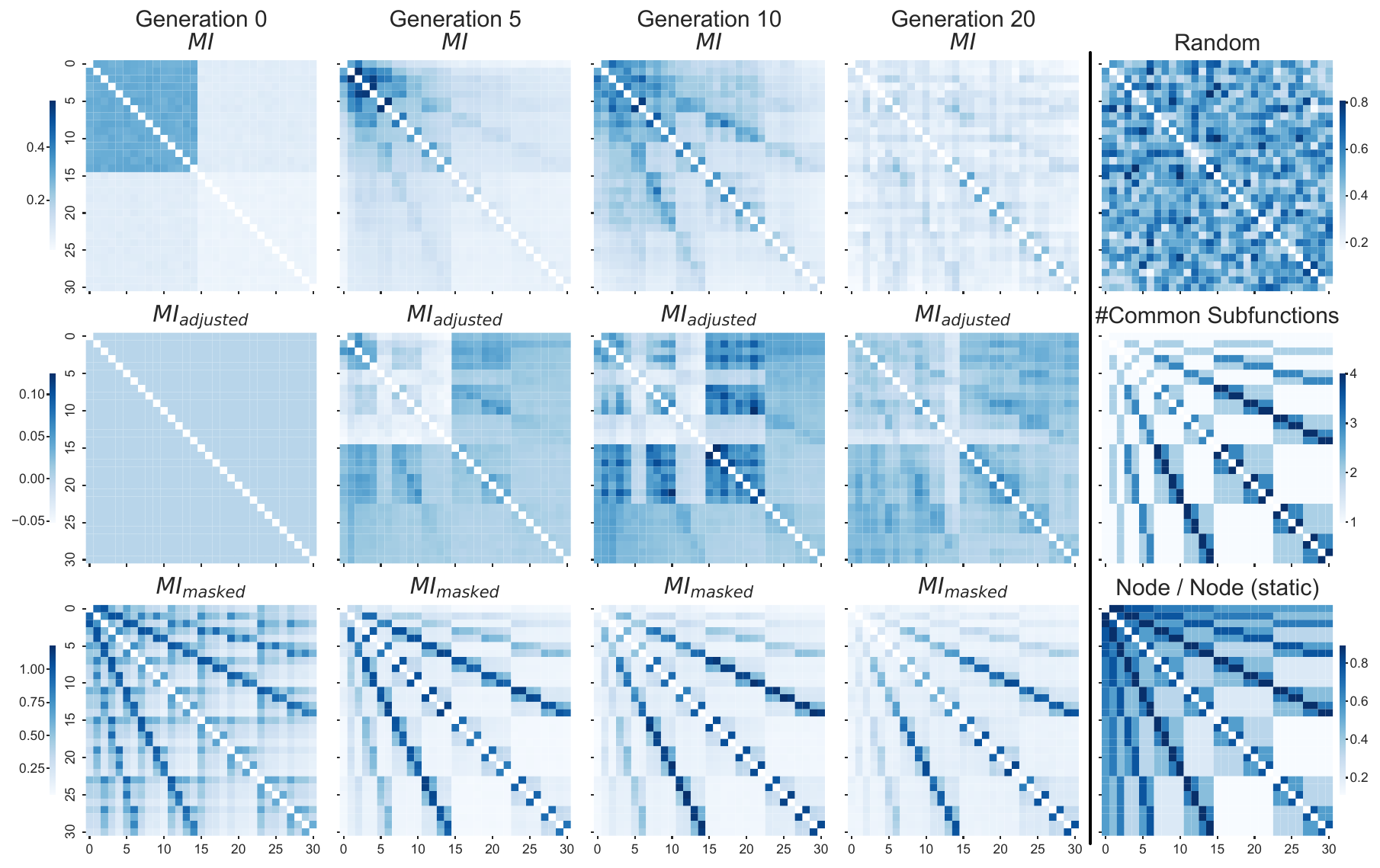}
        \caption{Concrete Compressive Strength}
    \end{subfigure}
    \begin{subfigure}[t]{0.495\linewidth}
        \centering
        \includegraphics[width=1.0\linewidth]{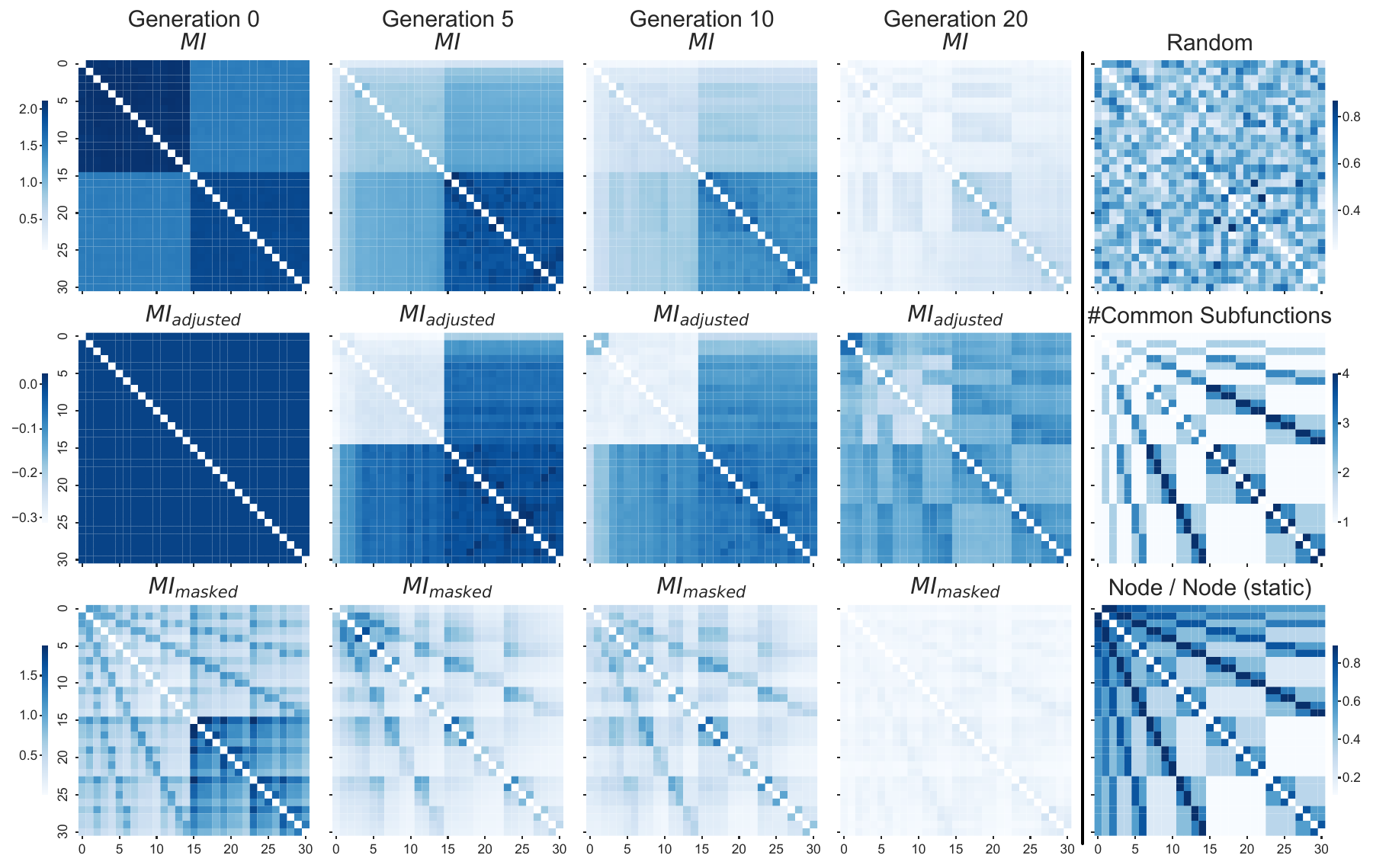}
        \caption{Dow Chemical}
    \end{subfigure}
    \hfill
    \begin{subfigure}[t]{0.495\linewidth}
        \centering
        \includegraphics[width=1.0\linewidth]{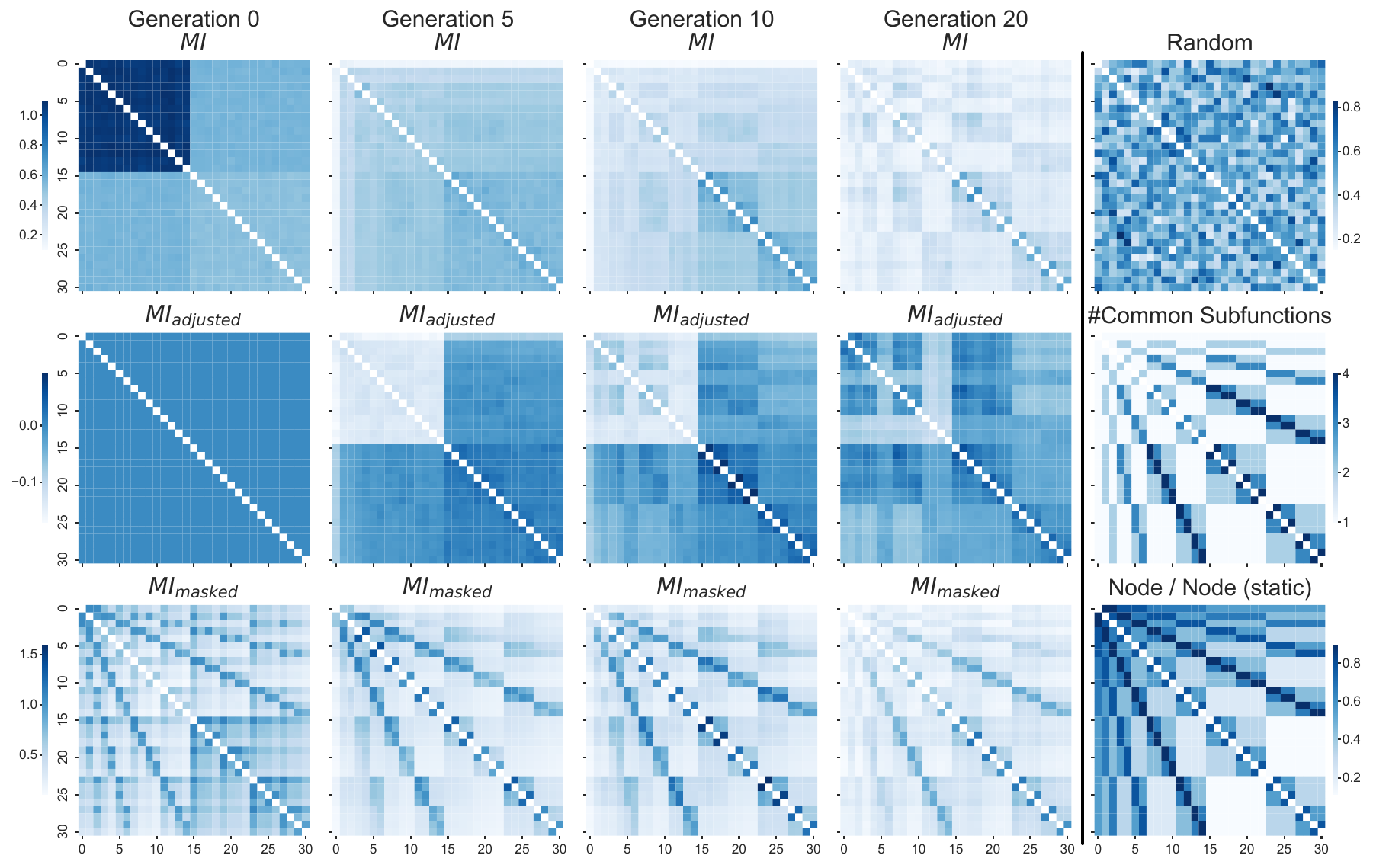}
        \caption{Tower}
    \end{subfigure}
    \caption{The different linkage measures over generations, averaged over 30 runs per problem with a template height of 5 (31 Nodes), LS and a fixed population size of 1024. Note that different scales are used for each measure as the focus lies on differences between variable pairs within measures instead of differences between linkage measures.}
    \label{fig:similarity_other}
\end{figure*}

\begin{figure*}[htbp]
    \centering
    \begin{subfigure}[t]{0.65\linewidth}
        \centering
        \includegraphics[width=1.0\linewidth]{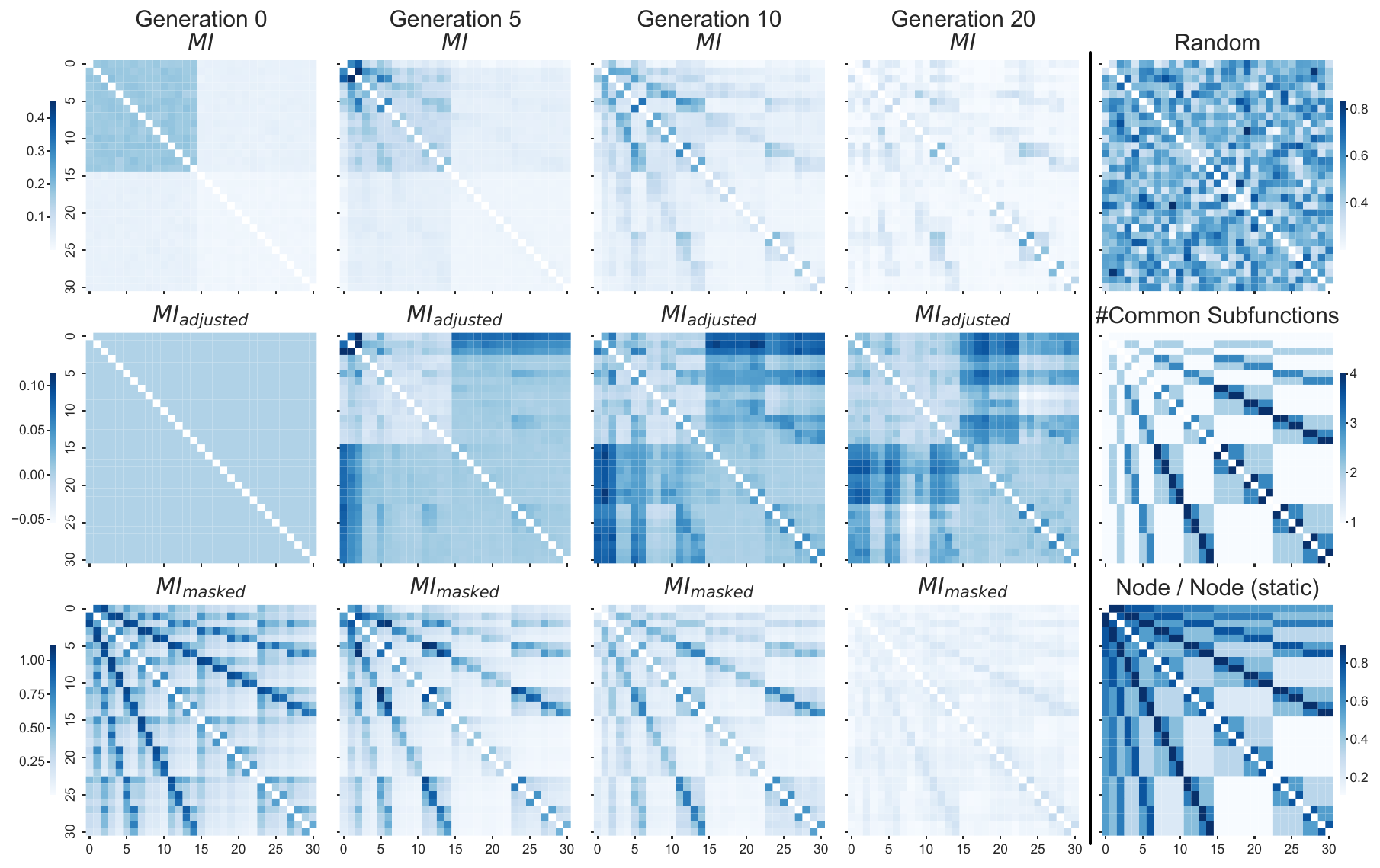}
        \Description{For generations 0 (i.e. after initialization and before variation), 5, 10 and 20, the averaged pairwise similarity measures are shown as heatmaps. Initially MI shows clear bias, while $MI_{adjusted}$ assumes no linkage between variables in the first generation. Both $MI_{masked}$ is very similar to Node, showing that the template structure is learned. With increasing generations, both MI and $MI_{adjusted}$ also increasingly resemble the Node measure, albeit less distinctively.}
        \caption{Airfoil}
    \end{subfigure}
    \begin{subfigure}[t]{0.495\linewidth}
        \centering
        \includegraphics[width=1.0\linewidth]{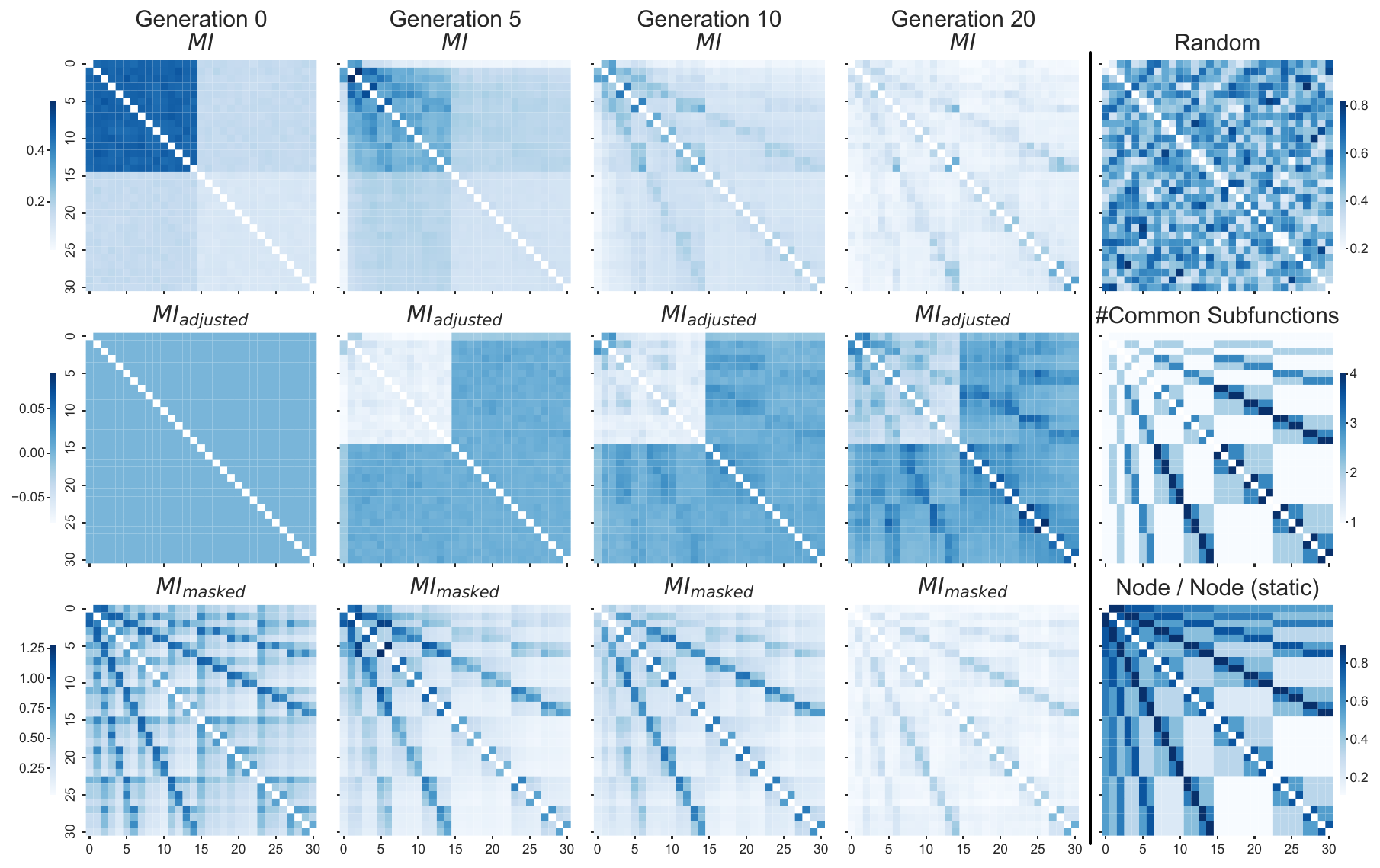}
        \Description{For generations 0 (i.e. after initialization and before variation), 5, 10 and 20, the averaged pairwise similarity measures are shown as heatmaps. Initially MI shows clear bias, while $MI_{adjusted}$ assumes no linkage between variables in the first generation. Both $MI_{masked}$ is very similar to Node, showing that the template structure is learned. With increasing generations, both MI and $MI_{adjusted}$ also increasingly resemble the Node measure, albeit less distinctively.}
        \caption{Bike Sharing (Daily)}
    \end{subfigure}
    \hfill
    \begin{subfigure}[t]{0.495\linewidth}
        \centering
        \includegraphics[width=1.0\linewidth]{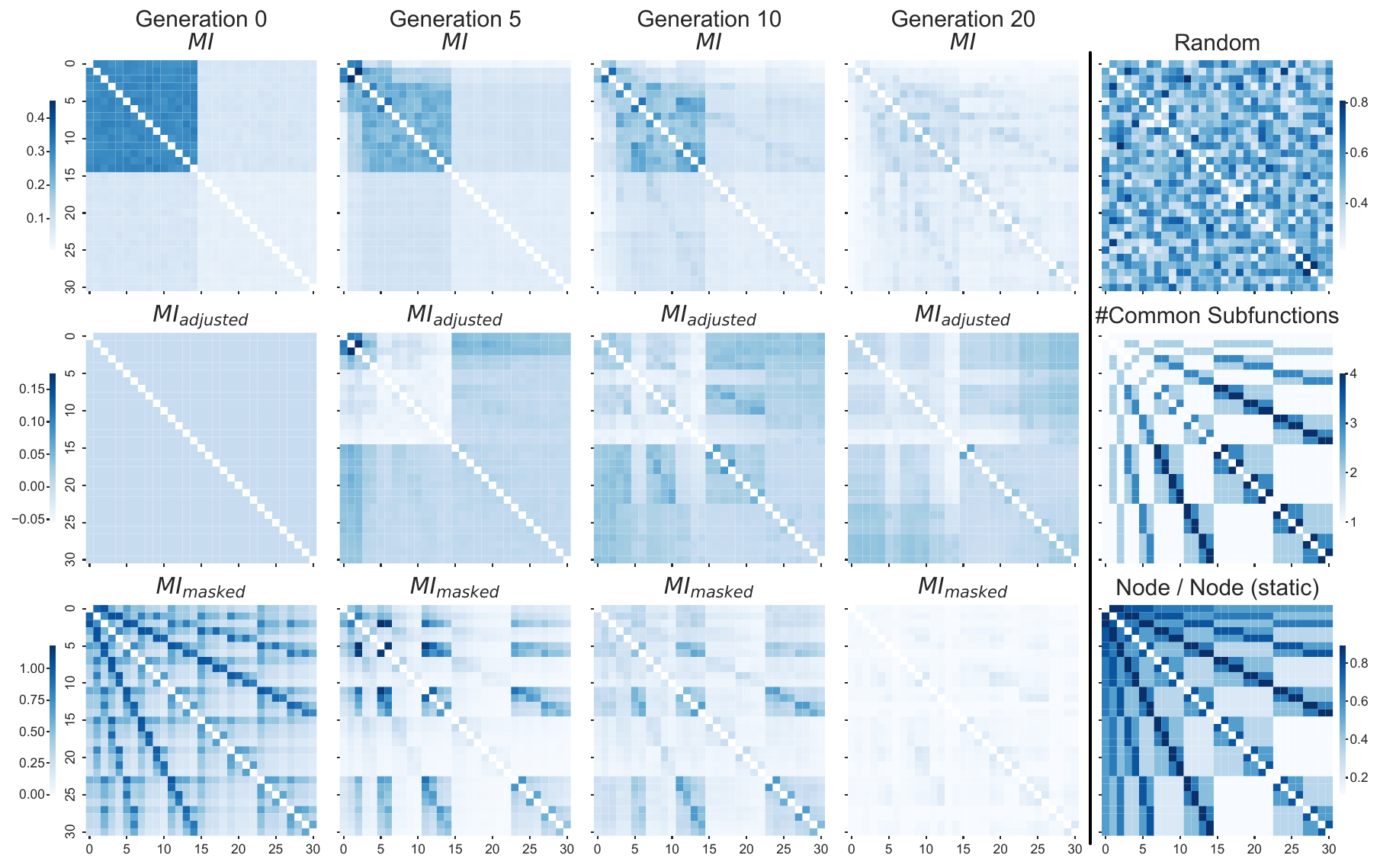}
        \Description{For generations 0 (i.e. after initialization and before variation), 5, 10 and 20, the averaged pairwise similarity measures are shown as heatmaps. Initially MI shows clear bias, while $MI_{adjusted}$ assumes no linkage between variables in the first generation. Both $MI_{masked}$ is very similar to Node, showing that the template structure is learned. With increasing generations, both MI and $MI_{adjusted}$ also increasingly resemble the Node measure, albeit less distinctively.}
        \caption{Concrete Compressive Strength}
    \end{subfigure}
    \begin{subfigure}[t]{0.495\linewidth}
        \centering
        \includegraphics[width=1.0\linewidth]{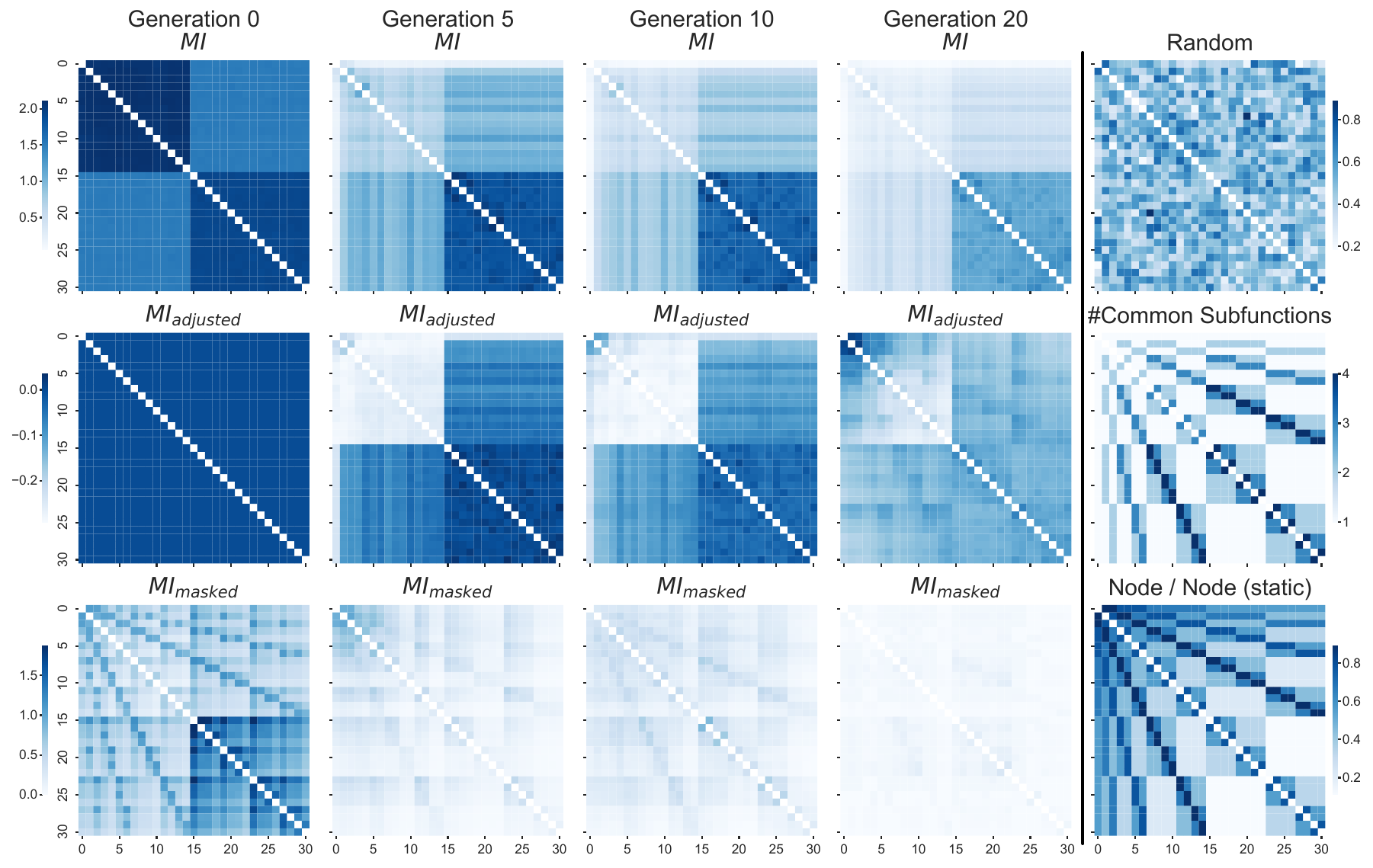}
        \Description{For generations 0 (i.e. after initialization and before variation), 5, 10 and 20, the averaged pairwise similarity measures are shown as heatmaps. Initially MI shows clear bias, while $MI_{adjusted}$ assumes no linkage between variables in the first generation. Both $MI_{masked}$ is very similar to Node, showing that the template structure is learned. With increasing generations, both MI and $MI_{adjusted}$ also increasingly resemble the Node measure, albeit less distinctively.}
        \caption{Dow Chemical}
    \end{subfigure}
    \hfill
    \begin{subfigure}[t]{0.495\linewidth}
        \centering
        \includegraphics[width=1.0\linewidth]{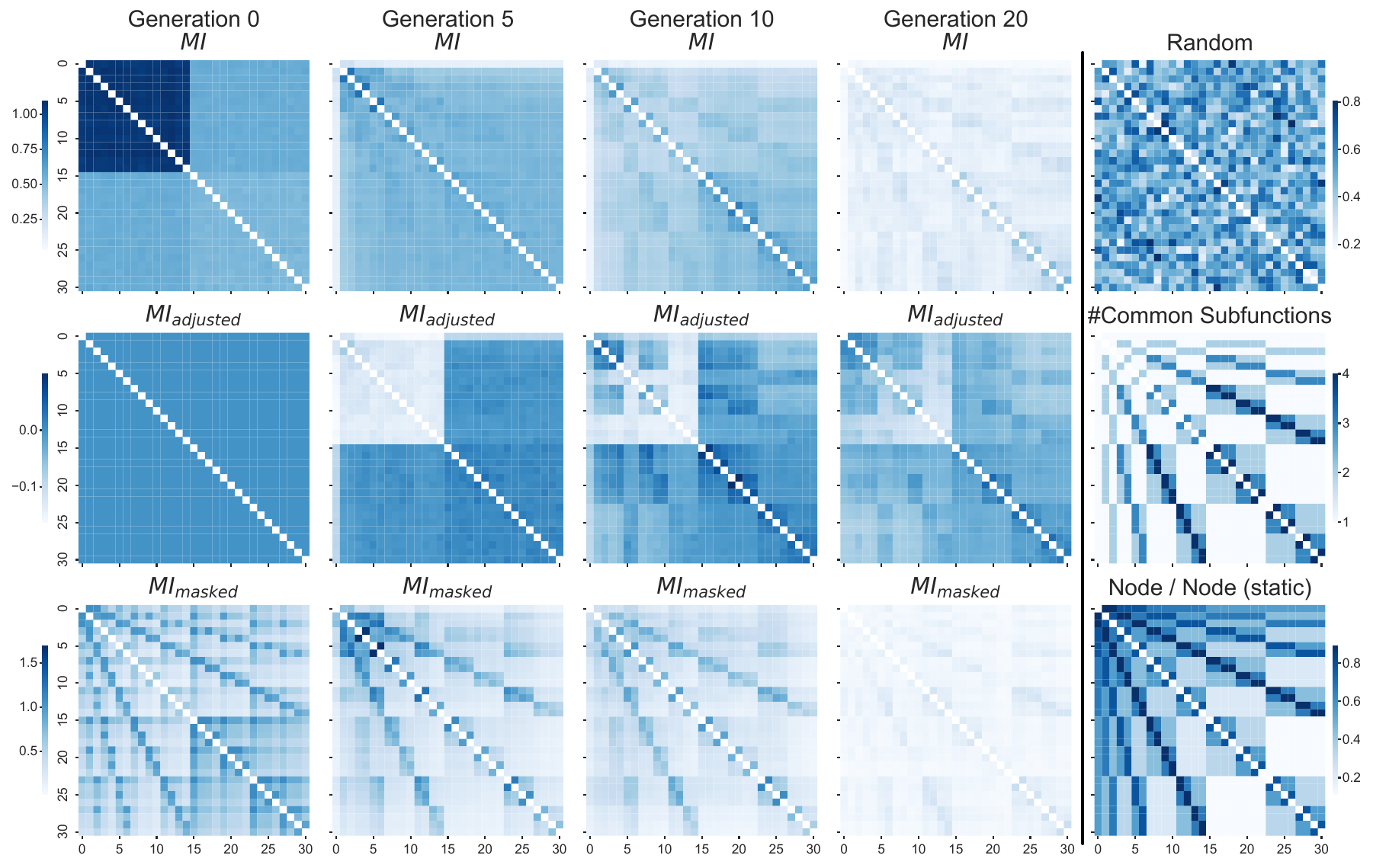}
        \Description{For generations 0 (i.e. after initialization and before variation), 5, 10 and 20, the averaged pairwise similarity measures are shown as heatmaps. Initially MI shows clear bias, while $MI_{adjusted}$ assumes no linkage between variables in the first generation. Both $MI_{masked}$ is very similar to Node, showing that the template structure is learned. With increasing generations, both MI and $MI_{adjusted}$ also increasingly resemble the Node measure, albeit less distinctively.}
        \caption{Tower}
    \end{subfigure}
    \caption{The different linkage measures over generations, averaged over 30 runs per problem with a template height of 5 (31 Nodes), \emph{without} LS and a fixed population size of 1024. Note that different scales are used for each measure as the focus lies on differences between variable pairs within measures instead of differences between linkage measures.}
    \label{fig:similarity_no_ls}
\end{figure*}

\begin{figure*}[htbp]
    \centering
    \begin{subfigure}[t]{0.65\linewidth}
        \centering
        \includegraphics[width=1.0\linewidth]{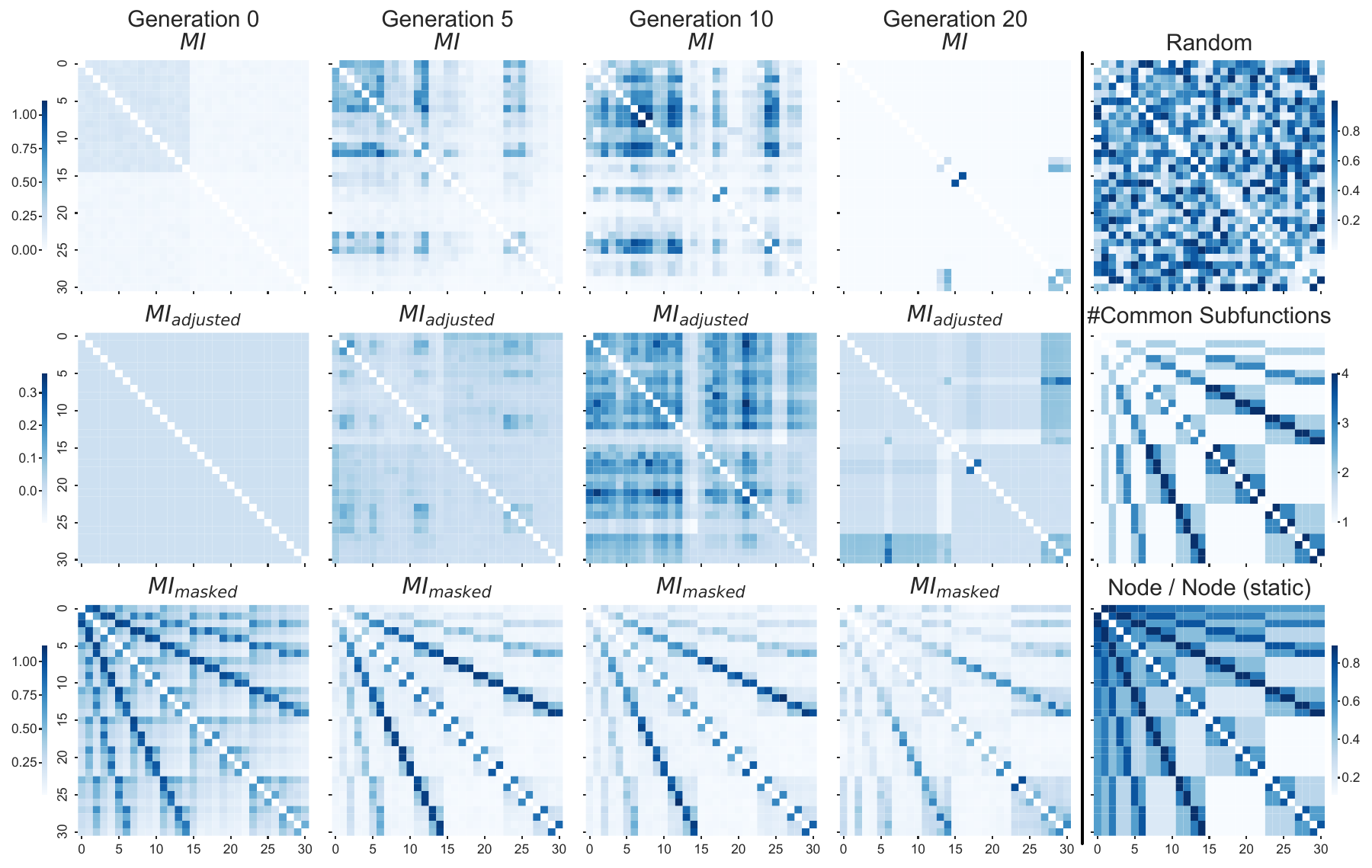}
        \Description{For generations 0 (i.e. after initialization and before variation), 5, 10 and 20, the averaged pairwise similarity measures are shown as heatmaps. Initially MI shows clear bias, while $MI_{adjusted}$ assumes no linkage between variables in the first generation. Both $MI_{masked}$ is very similar to Node, showing that the template structure is learned. With increasing generations, both MI and $MI_{adjusted}$ also increasingly resemble the Node measure, albeit less distinctively.}
        \caption{Airfoil}
    \end{subfigure}
    \begin{subfigure}[t]{0.495\linewidth}
        \centering
        \includegraphics[width=1.0\linewidth]{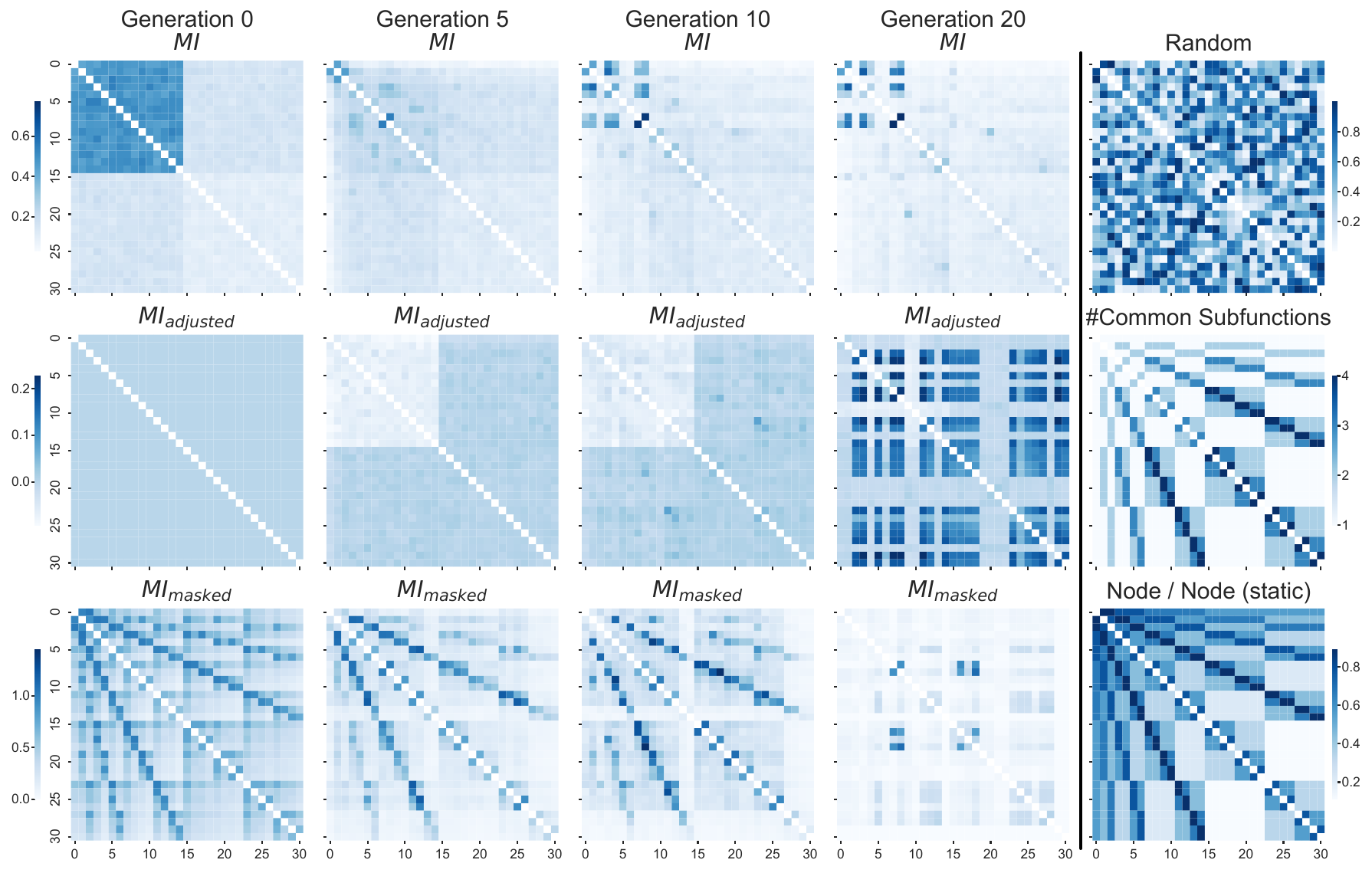}
        \Description{For generations 0 (i.e. after initialization and before variation), 5, 10 and 20, the averaged pairwise similarity measures are shown as heatmaps. Initially MI shows clear bias, while $MI_{adjusted}$ assumes no linkage between variables in the first generation. Both $MI_{masked}$ is very similar to Node, showing that the template structure is learned. With increasing generations, both MI and $MI_{adjusted}$ also increasingly resemble the Node measure, albeit less distinctively.}
        \caption{Bike Sharing (Daily)}
    \end{subfigure}
    \hfill
    \begin{subfigure}[t]{0.495\linewidth}
        \centering
        \includegraphics[width=1.0\linewidth]{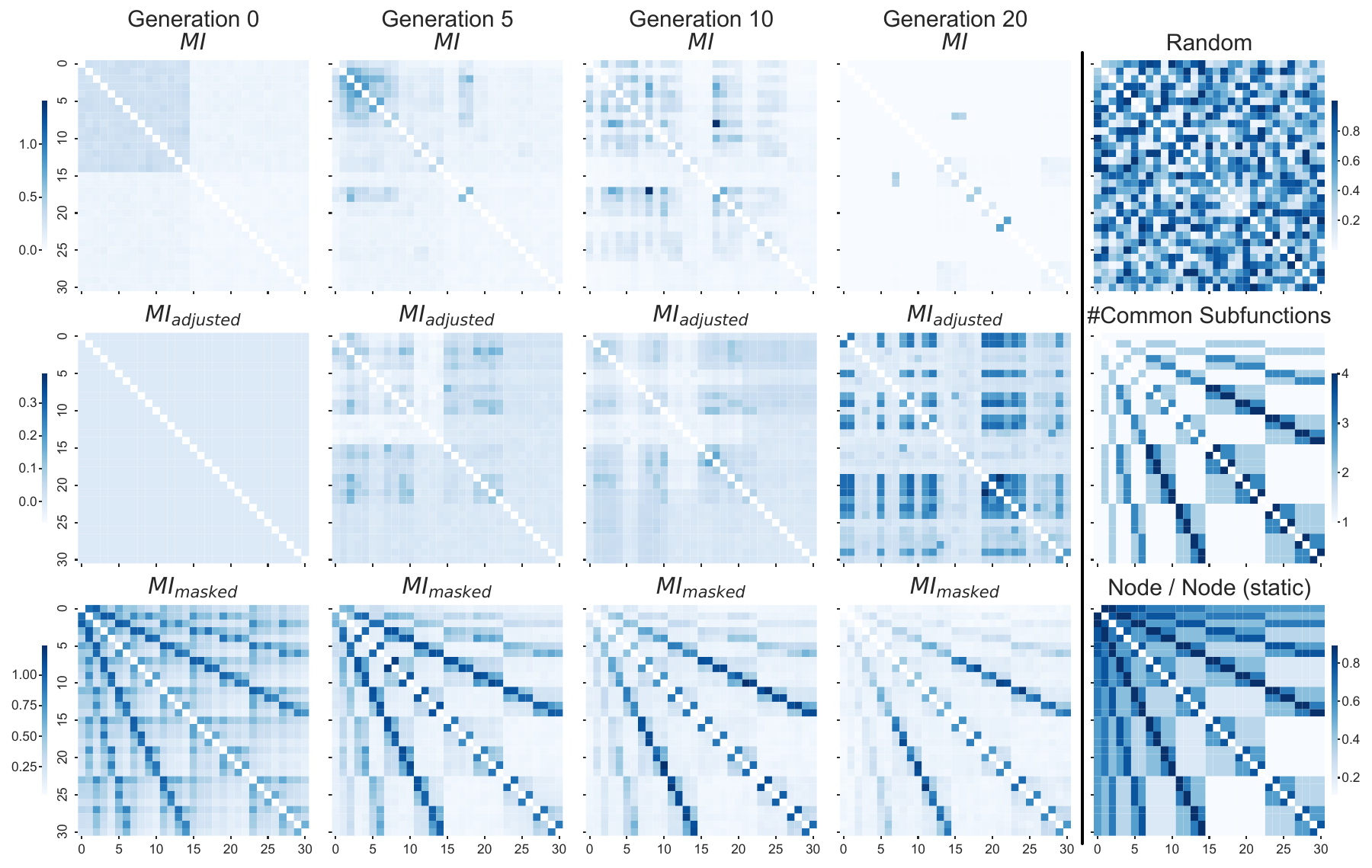}
        \Description{For generations 0 (i.e. after initialization and before variation), 5, 10 and 20, the averaged pairwise similarity measures are shown as heatmaps. Initially MI shows clear bias, while $MI_{adjusted}$ assumes no linkage between variables in the first generation. Both $MI_{masked}$ is very similar to Node, showing that the template structure is learned. With increasing generations, both MI and $MI_{adjusted}$ also increasingly resemble the Node measure, albeit less distinctively.}
        \caption{Concrete Compressive Strength}
    \end{subfigure}
    \begin{subfigure}[t]{0.495\linewidth}
        \centering
        \includegraphics[width=1.0\linewidth]{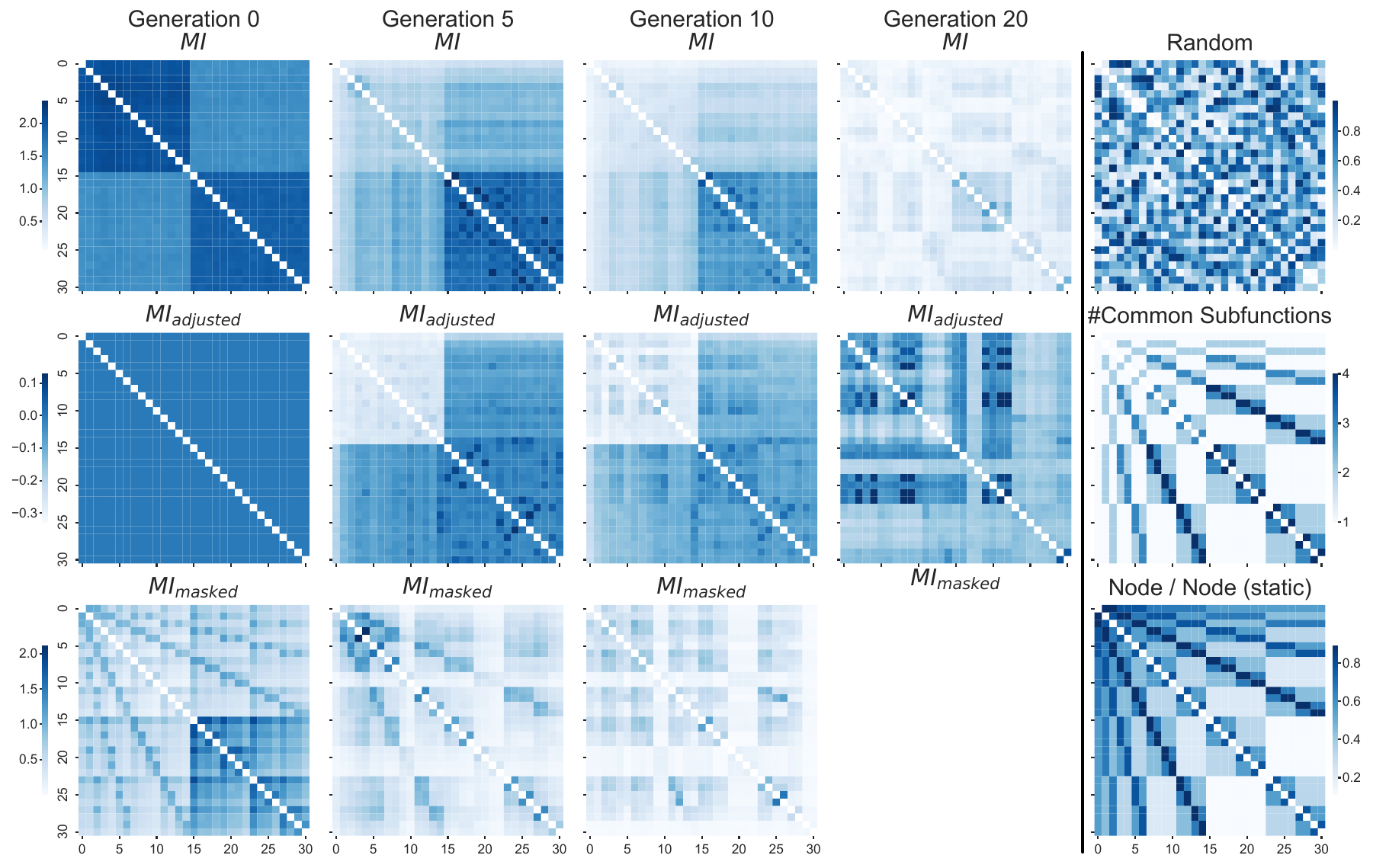}
        \Description{For generations 0 (i.e. after initialization and before variation), 5, 10 and 20, the averaged pairwise similarity measures are shown as heatmaps. Initially MI shows clear bias, while $MI_{adjusted}$ assumes no linkage between variables in the first generation. Both $MI_{masked}$ is very similar to Node, showing that the template structure is learned. With increasing generations, both MI and $MI_{adjusted}$ also increasingly resemble the Node measure, albeit less distinctively.}
        \caption{Dow Chemical}
    \end{subfigure}
    \hfill
    \begin{subfigure}[t]{0.495\linewidth}
        \centering
        \includegraphics[width=1.0\linewidth]{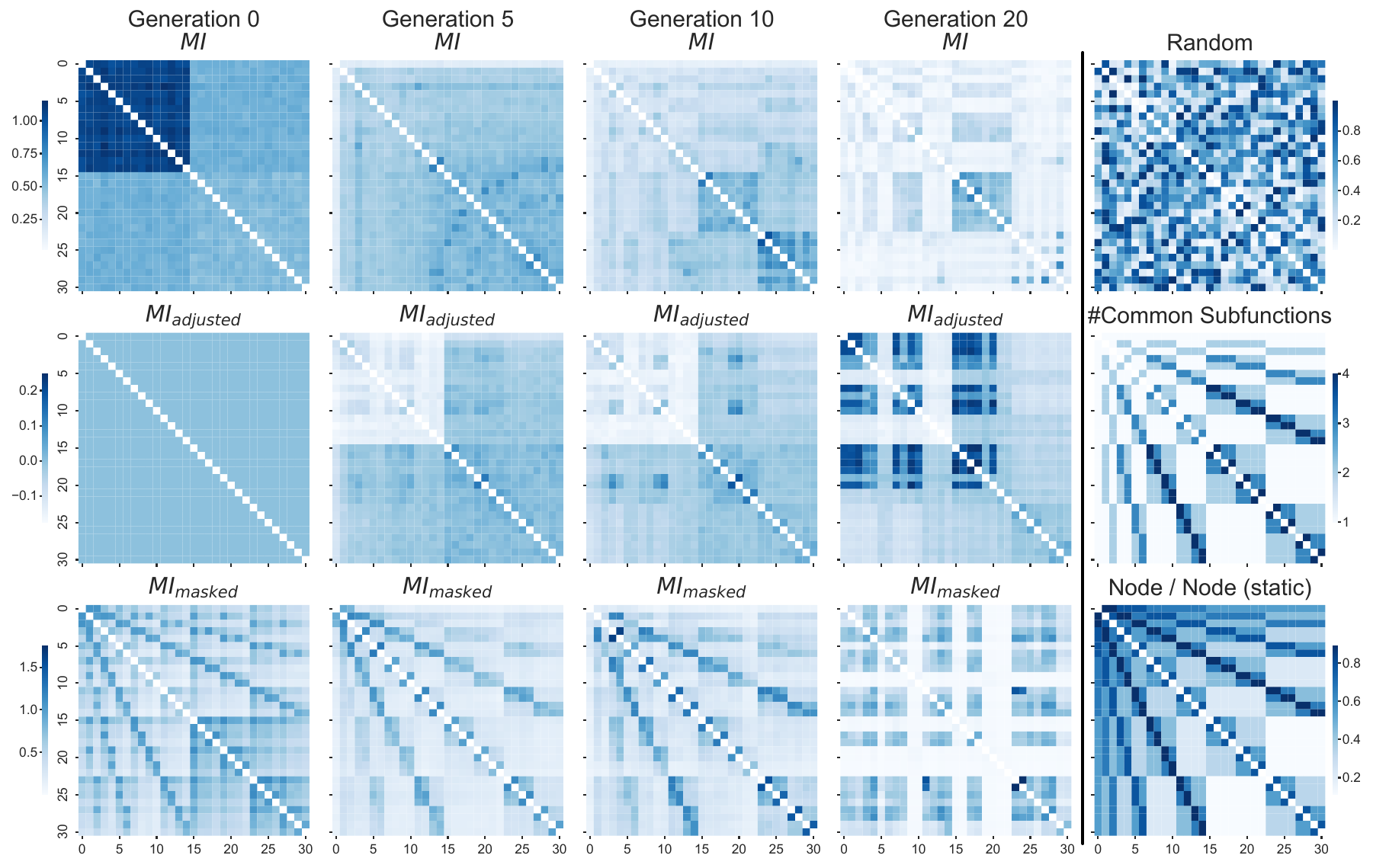}
        \Description{For generations 0 (i.e. after initialization and before variation), 5, 10 and 20, the averaged pairwise similarity measures are shown as heatmaps. Initially MI shows clear bias, while $MI_{adjusted}$ assumes no linkage between variables in the first generation. Both $MI_{masked}$ is very similar to Node, showing that the template structure is learned. With increasing generations, both MI and $MI_{adjusted}$ also increasingly resemble the Node measure, albeit less distinctively.}
        \caption{Tower}
    \end{subfigure}
    \caption{The different linkage measures over generations, for the first of 30 runs per problem with a template height of 5 (31 Nodes), LS and a fixed population size of 1024. Note that different scales are used for each measure, as the focus lies on differences between variable pairs within measures instead of differences between linkage measures. Empty matrices indicate that the run finished before reaching the indicated generation.}
    \label{fig:similarity_runs_ls}
\end{figure*}

To assess whether the linkage structure learned corresponds to the structural template across the problems considered,~\Cref{fig:similarity_other} displays the averaged linkage across runs with linear scaling, and ~\Cref{fig:similarity_no_ls} shows the results from an additional repeat of the experiment without linear scaling. Without linear scaling, the averaged structures still generally align with the node proximity and the initial bias of MI is clearly visible. However, the emerging structure tends to be noisier compared to with LS enabled, in particular for the Dow Chemical dataset. In addition to the pairwise similarity shown when averaged across all 30 runs, single run results with LS enabled are shown in~\Cref{fig:similarity_runs_ls} to give some indication of the variance between runs. Again, the masked MI variant approximates the template structure and MI shows an initial bias.

\section{Effect of Operator Set Size\label{appendix:operators}}

\begin{figure*}[t]
    \centering
    \includegraphics[width=0.95\linewidth]{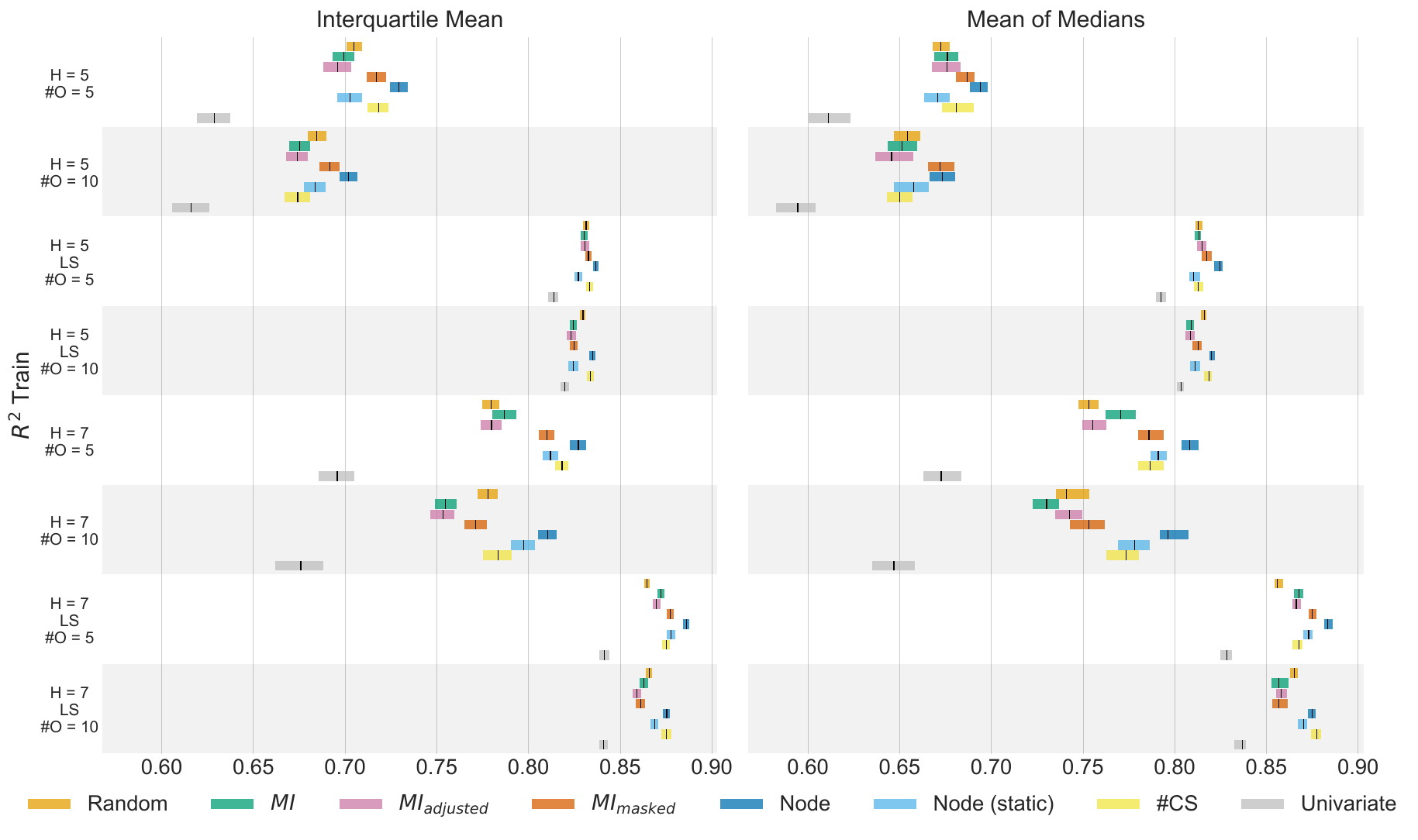}
    \vspace{-0.5cm}
    \Description{The figure shows the aggregate $R^2$ scores (higher is better) on the problems for each combination of template height, LS and operator sets considered with confidence intervals. For each combination, both for the training and testing accuracy, there tends to be one group of less accurate measures containing Random, $MI_{adjusted}$ and MI (mostly in this order with improving accuracy). $MI_{masked}$ tends to perform clearly better, with Node generally outperforming all other methods. The static version of Node performs worse compared to node, sometimes better than the group of existing methods and sometimes comparably. With linear scaling enabled all methods are closer together, and the differences between methods increase with the larger templates using height 7. Performance with the larger operator set is worse across all settings.}
    \caption{Aggregate $R^2$ scores (higher is better) on the problems for each combination of template height, LS and operator set size considered in the extended experiment. The interquartile mean corresponds to the mean after discarding the bottom and top 25\% of runs for each problem, and the mean of medians corresponds to the mean of the median performances on each problem. The colored bar corresponds to the 95\% confidence interval estimated using a percentile bootstrap with stratified sampling as per~\cite{agarwalDeepReinforcementLearning} with the expected value in black.}
    \label{fig:operators}
\end{figure*}

\begin{figure*}[tbp]
    \centering
    \includegraphics[width=\linewidth]{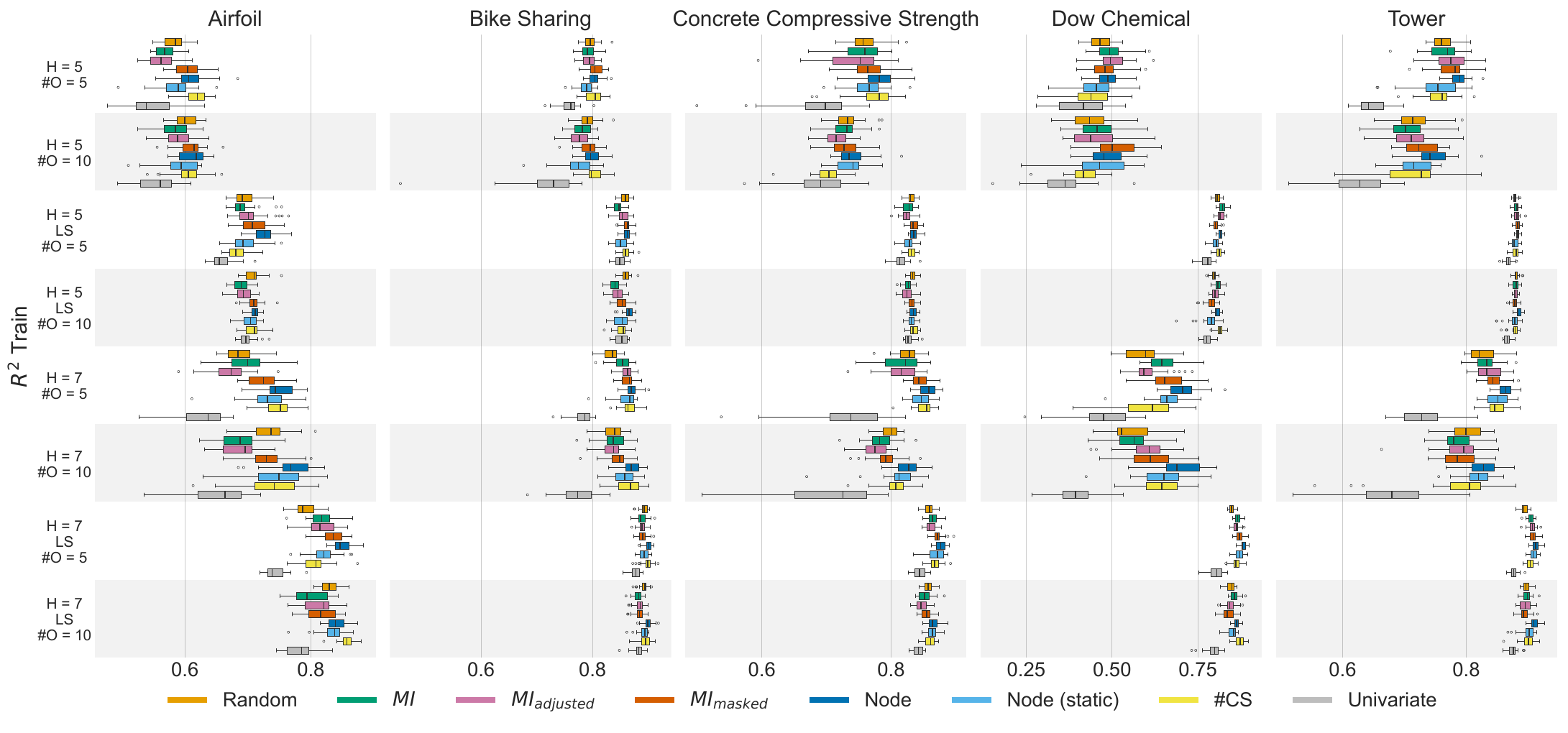}
    \vspace{-0.8cm}
    \Description{Final results for all problems and settings considered (including the varied operator set size), showcasing the training accuracy. Generally, Node performs best, followed by $MI_{masked}$, Common Subfunction Count, Node (static), $MI_{adjusted}$/MI, Random and Univariate in roughly that order across problems.}
    \caption{$R^2$ score (higher is better) across problem, template, linear scaling and operator set combination considered. The boxes show the quartiles and the whiskers extend to points that lie within 1.5 inter-quartile ranges of the lower and upper quartile. Observations outside this range are displayed independently.}
    \label{fig:operators_per_problem}
\end{figure*}

The first experiment in the paper considers the combinations of template height and linear scaling with a single operator set. However, the operator set itself is another core parameter that potentially affects linkage and thus the first experiment was repeated with a larger operator set containing \(\{+,-,\times,\div,\sin,\cos,\exp,\log,\sqrt{\cdot},\cdot^2\}\). The interval estimate comparisons factorized by setting for this larger experiment are shown in~\Cref{fig:operators} and~\Cref{fig:operators_per_problem} shows per problem results. The additional operators tend to perform worse compared to the smaller operator set, while the relative performance between the various linkage measures remains largely unaffected. Notably, the mutual information based measures tend to perform worse compared to the random measure when the larger operator set is used.

\end{document}